%% file: neurips_2026.tex
\documentclass{article}

\PassOptionsToPackage{numbers, compress}{natbib}
\usepackage[colorlinks=true, citecolor=cyan, linkcolor=black]{hyperref}
\usepackage[preprint]{neurips_2026}

\usepackage[utf8]{inputenc}
\usepackage[T1]{fontenc}
\usepackage{hyperref}
\usepackage{url}
\usepackage{booktabs}
\usepackage{amsfonts}
\usepackage{amsmath}
\usepackage{nicefrac}
\usepackage{microtype}
\usepackage{xcolor}
\usepackage{graphicx}
\usepackage{enumitem}
\usepackage{xspace}
\usepackage{pifont}
\usepackage{multirow}
\usepackage{colortbl}
\usepackage{makecell}
\usepackage{subcaption}
\usepackage{listings}
\usepackage{algorithm}
\usepackage{tikz}

\definecolor{ilrow}{HTML}{E8F1FB}
\definecolor{vlarow}{HTML}{E8F5E9}
\definecolor{wamrow}{HTML}{FFF3E0}
\definecolor{layerone}{HTML}{ECEFF1}
\definecolor{layertwo}{HTML}{E3F2FD}
\definecolor{layerthree}{HTML}{E8F5E9}
\definecolor{layerfour}{HTML}{FFF3E0}
\definecolor{layerfive}{HTML}{F3E5F5}
\definecolor{layersix}{HTML}{FCE4EC}
\definecolor{layeronefg}{HTML}{455A64}
\definecolor{layertwofg}{HTML}{1565C0}
\definecolor{layerthreefg}{HTML}{2E7D32}
\definecolor{layerfourfg}{HTML}{EF6C00}
\definecolor{layerfivefg}{HTML}{7B1FA2}
\definecolor{layersixfg}{HTML}{C2185B}
\definecolor{nautiluscmd}{HTML}{005A9C}

\definecolor{codekw}{HTML}{1F4E79}   
\definecolor{codefn}{HTML}{6F42C1}   
\definecolor{codecm}{HTML}{6A737D}   
\definecolor{codestr}{HTML}{0B6623}  
\lstdefinestyle{nautilus}{%
  language=Python,
  basicstyle=\scriptsize\ttfamily,
  keywordstyle=\color{codekw}\bfseries,
  commentstyle=\color{codecm}\itshape,
  stringstyle=\color{codestr},
  emph={%
    BasePolicy, WebsocketPolicyServer, WebsocketClientPolicy,
    ActionChunkBroker, MyPolicy, infer, reset, serve_forever,
    get_server_metadata, load_model, preprocess, make_env,
    Policy, Benchmark, Robot, Observation, Action, Transition,
    step, get_observation, apply_action, safe_stop%
  },
  emphstyle=\color{codefn}\bfseries,
  showstringspaces=false,
  columns=fullflexible,
  keepspaces=true,
  upquote=true,
  aboveskip=2pt, belowskip=2pt,
  xleftmargin=0pt, framexleftmargin=0pt,
}

\newcommand{\cmark}{\textcolor{green!55!black}{\ding{51}}}
\newcommand{\xmark}{\textcolor{red!75!black}{\ding{55}}}
\newcommand{\pmark}{\textcolor{orange!85!black}{$\circ$}}

\newcommand{\layertag}[2]{\textcolor{#1}{\textbf{#2}}}
\newcommand{\Lone}{\layertag{layeronefg}{L1}}
\newcommand{\Ltwo}{\layertag{layertwofg}{L2}}
\newcommand{\Lthree}{\layertag{layerthreefg}{L3}}
\newcommand{\Lfour}{\layertag{layerfourfg}{L4}}
\newcommand{\Lfive}{\layertag{layerfivefg}{L5}}
\newcommand{\Lsix}{\layertag{layersixfg}{L6}}
\newcommand{\nautcmd}[1]{%
  \textcolor{nautiluscmd}{\texttt{> #1}}%
}

\newcommand{\sysname}{\textsc{Nautilus}\xspace}

\title{Nautilus: From One Prompt to Plug-and-Play \\Robot Learning}

\author{\mdseries%
  Yufeng Jin\textsuperscript{1}\thanks{Equal contribution.} \and
  Jianfei Guo\textsuperscript{6}\footnotemark[1] \and
  Xiaogang Jia\textsuperscript{2} \and
  Yu Deng\textsuperscript{1} \and
  Zechu Li\textsuperscript{1} \and
  Han Liu\textsuperscript{6} \and
  Weiran Liao\textsuperscript{2} \and
  Vignesh Prasad\textsuperscript{1} \and
  Mathias Franzius\textsuperscript{7} \and
  Gerhard Neumann\textsuperscript{2,3} \and
  Georgia Chalvatzaki\textsuperscript{1,4,5}%
  \thanks{%
    \textsuperscript{1}TU Darmstadt;\,
    \textsuperscript{2}KIT;\,
    \textsuperscript{3}FZI;\,
    \textsuperscript{4}Hessian.AI;\,
    \textsuperscript{5}Robotics Institute Germany;\,
    \textsuperscript{6}Independent Researcher;\,
    \textsuperscript{7}Honda Research Institute Europe.%
  }%
}

\begin{document}

\maketitle

\input{sections/00_abstract.tex}
\begin{figure}[t!]
  \setlength{\abovecaptionskip}{2pt}
  \setlength{\belowcaptionskip}{0pt}
  \centering
  \includegraphics[width=\linewidth]{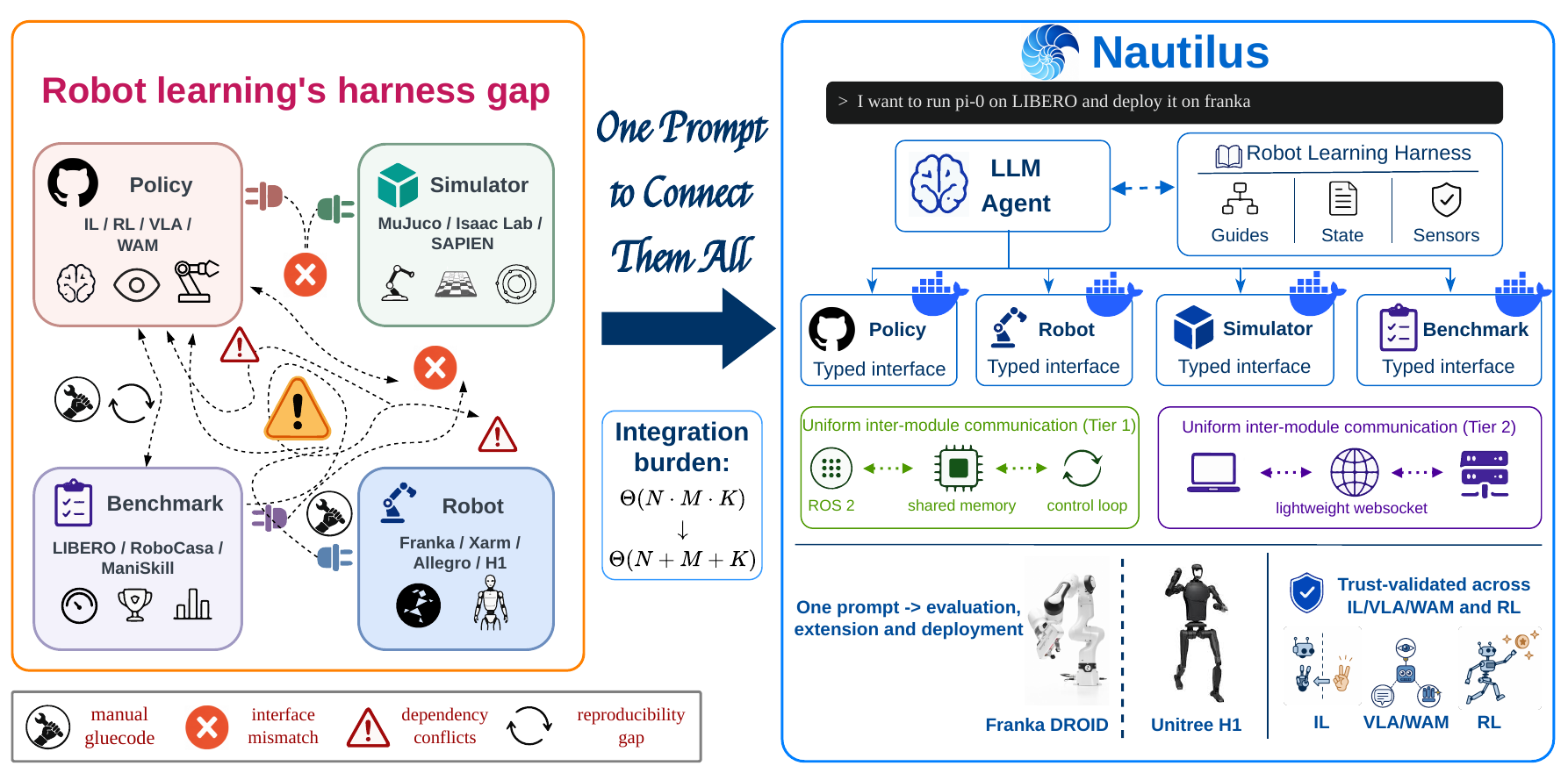}
  \caption{\footnotesize\textbf{The harness gap in robot learning
  research, and how \sysname{} closes it.}
  Existing workflows repeatedly rebuild policy--benchmark--robot glue,
  yielding a field-scale integration burden of
  $\Theta(N\!\cdot\!M\!\cdot\!K)$. \sysname{} introduces a shared
  robot learning harness: a substrate of typed contracts, chambered
  execution, and uniform transport, plus a content layer of Guides,
  Sensors, and State. This changes the unit of work from pairwise
  integration to reusable onboarding, reducing the burden to
  $\Theta(N\!+\!M\!+\!K)$.}
  \label{fig:teaser}
\end{figure}

\input{sections/01_introduction.tex}

\input{sections/02_related_work.tex}
\input{sections/03_method.tex}
\input{sections/04_experiments.tex}
\input{sections/05_discussion.tex}
\input{sections/06_conclusion.tex}

\newpage
\bibliographystyle{unsrtnat}
\bibliography{references}

\newpage
\appendix
\input{sections/A_appendix.tex}


\end{document}

%% file: sections/00_abstract.tex
\begin{abstract}
Robot learning research is fragmented across policy families, benchmark suites, and real robots;
Each implementation is entangled with the others in a complex combination matrix, making it an engineering nightmare to port any single element.
General-purpose coding agents may occasionally bridge specific setups, but cannot close this gap at scale because they lack the procedural priors and validation practices that characterize robotics research workflows.
We propose \sysname{}, an open-source harness that turns a single user prompt —for example, ``Evaluate policy A with benchmark B'' — into ready-to-use reproduction, evaluation, fine-tuning, and deployment workflows.
\sysname{} provides:
plug-and-play agent skill sets with distilled priors from robotics research;
typed contracts among policies, simulators/benchmarks, and real-world robots;
unified interfaces and execution environments;
and a trustworthy agentic coding workflow with explicit, automated validation, and testing at each milestone. 
\sysname{} can not only automatically generate the required adapters and containers for existing implementations, but also wrap and onboard new or user-provided policies, simulators/benchmarks, and robots, all connected via a uniform interface. This expands cross-validation coverage without hand-written glue code.
Like a nautilus shell that grows by adding chambers, \sysname{} scales by extending its execution in chambered units, making it a research harness for scalability rather than a hand-curated framework, and aiming to reduce the engineering burden of cross-family reproduction and evaluation in the ever-growing robot learning ecosystem.
Project website: \url{https://yufengjin.github.io/nautilus/}.
\end{abstract}

%% file: sections/01_introduction.tex
\section{Introduction}
\label{sec:intro}
Large Language Models (LLMs) are now widely used in code
generation and software-engineering workflows~\citep{jiang2026survey, dong2025survey}.
In robotics, LLMs have already been explored in task-level embodied
pipelines, including policy code generation and reward
design~\citep{liang2023code, ma2024eureka}. Our setting is different: we study agentic coding systems as orchestrators of robot learning research workflows across repositories, simulators, benchmarks, and hardware interfaces, where even strong coding agents still often fail to reach an end-to-end reproduced result~\citep{meng2026agentharness, bui2026buildingeffectiveaicoding}.
We argue that this persistent failure is not primarily a
model-capability problem, but a \textbf{harness problem}. \emph{Harness
engineering}---the design of everything around an LLM except the model
itself, including orchestration, tools, verification protocols,
guardrails, state, and procedural priors---has recently been identified
as a central determinant of agentic project
success~\citep{meng2026agentharness, bui2026buildingeffectiveaicoding, pan2026naturallanguageagentharnesses, lee2026metaharnessendtoendoptimizationmodel}.
Robot learning still lacks a general-purpose research harness.

Robot learning research
spans three largely independent axes: policy families $P$
(e.g., vision-language-action models (VLAs), imitation learning (IL),
reinforcement learning (RL), and world action model (WAM)), benchmark suites $B$
(e.g., LIBERO~\citep{liu2023libero},
RoboCasa~\citep{nasiriany2024robocasa},
ManiSkill~\citep{tao2024maniskill3},
ALOHA~\citep{zhao2023learning}, etc.), and robot embodiments $R$
(e.g., single-arm, bimanual, dexterous-hand, locomotion, humanoid,
etc.), with cardinalities $N$, $M$, and $K$, respectively. Each
non-trivial $(P, B, R)$ cross-comparison typically requires a distinct
hand-written integration layer, so docker container setup, observation
adapters, smoke tests, and trust-validation procedures are repeatedly
re-implemented across papers, labs, and reproduction efforts. At the
field level, this induces a harness cost of
$\Theta(N\!\cdot\!M\!\cdot\!K)$, whereas a shared harness would
amortize the same burden to $\Theta(N\!+\!M\!+\!K)$.
Existing systems address only part of the problem.
Robot learning
frameworks such as LeRobot~\citep{lerobot}, Isaac
Lab~\citep{isaaclab}, and robosuite~\citep{robosuite} consolidate common
workflows but remain coupled to fixed policies and benchmarks, while
general-purpose coding agents such as
Claude Code~\citep{anthropic2025claudecode},
OpenHands~\citep{wang2025openhands}, and
SWE-agent~\citep{yang2024sweagent} provide the agent loop but not the
robotics-specific procedural knowledge these workflows require.

We propose \textsc{Nautilus}, which to our knowledge is the first
domain-specific harness for robot learning research. Unlike a
conventional robot learning framework, \textsc{Nautilus} is an
agent-facing harness over typed policy, benchmark, and robot modules,
while the underlying coding-agent loop is treated as replaceable.
\textsc{Nautilus} is organized into two layers. The first is a
\emph{substrate} built on three engineering invariants---typed interface
contracts, chambered execution, and a uniform inter-module transport.
This substrate is a prerequisite: without typed contracts and chambered
isolation, the agent has no stable surface to reason over and must
relearn per-project glue code for each new $(P, B, R)$ triplet. On top of this substrate sits a \emph{content
layer} that instantiates the Guides+Sensors framework of harness
engineering~\citep{meng2026agentharness} and extends it with State. In
our formulation, \textbf{Guides} shape the agent's behavior before code
generation, \textbf{Sensors} validate its outputs after generation, and
\textbf{State}---an MCP-served, JSON-schema-validated contract
registry---mediates between modules, reducing $(P\!\times\!B)$
integration cost from $\Theta(N\!\cdot\!M)$ to $\Theta(N\!+\!M)$. Figure~\ref{fig:teaser} illustrates the harness gap in robot learning
and how \textsc{Nautilus} closes it.

Our contributions are fourfold. \textbf{1)} We introduce a modular
substrate based on typed interface contracts, chambered execution, and
uniform inter-module transport, collapsing
$(P\!\times\!B\!\times\!R)$ integration cost from
$\Theta(N\!\cdot\!M\!\cdot\!K)$ to $\Theta(N\!+\!M\!+\!K)$ and giving
the agent a stable, policy-agnostic surface to reason over. \textbf{2)}
We distill the procedural priors of robotics research into reusable
agent skills, turning tacit workflow knowledge into explicit guidance.
\textbf{3)} We release \textsc{Nautilus} together with structured
agentic workflows and evaluation tools for reproduction,
cross-evaluation, finetuning, and deployment across four policy
families (IL, VLA, RL, and WAM), six benchmark suites, and four
embodiment classes; on real hardware, \texttt{nautilus-collect}
provides evidence that the same Robot contract carries across
embodiments, with deployment validated on Franka and Unitree H1.
\textbf{4)} We design \textsc{Nautilus} as an extensible community
platform where researchers can reproduce and evaluate policies or
benchmarks without modifying framework code, while new contributions enter the registry through 3--5 typed method contracts with
commit- and image-pinned reproducibility and trust validation.

%% file: sections/02_related_work.tex
\section{Related Work}
\label{sec:related}

\input{sections/Tabel_compare_platforms}

\noindent\textbf{LLM coding agents and harness engineering.}
Code-generating LLMs have evolved from prompt-centric methods to tool
use and, more recently, to full agent
loops~\citep{wei2022chain, yao2023react,
schick2023toolformer, yang2024sweagent,
anthropic2025claudecode, wang2025openhands}. As these systems have
matured, the main bottleneck has shifted from prompting to the
surrounding scaffolding, formalized as \emph{harness
engineering}~\citep{meng2026agentharness,
bui2026buildingeffectiveaicoding}. Recent work studies this layer
through abstractions such as Guides and Sensors and through unified
evaluation
scaffolds~\citep{meng2026agentharness, kapoor2026holistic,
bui2026buildingeffectiveaicoding,
pan2026naturallanguageagentharnesses,
lee2026metaharnessendtoendoptimizationmodel,
rabanser2026scienceaiagentreliability,
staufer20262025aiagentindex}. These systems provide strong
general-purpose harnesses, but not the robotics-specific priors needed
for robot learning workflows. \textsc{Nautilus} specializes this
harness perspective through typed interfaces, chambered execution, and
workflow-level validation.

\noindent\textbf{Robot learning paradigms.}
Robot learning spans several policy families. \emph{Imitation
learning} (IL) fits policies to expert
trajectories~\citep{zhao2023learning, florence2022implicit,
jia2026pointmappolicy, chi2025diffusion, fu2024mobile,
chi2024universal, Ze2024DP3, octo_2023, haldar2024baku, hou2025dita},
while \emph{reinforcement learning} (RL) optimizes behavior through
reward-driven interaction with the
environment~\citep{schulman2017proximal,
haarnoja2018softactorcriticoffpolicymaximum, fujimoto2018addressing,
stable-baselines3, makoviychuk2021isaac}. More recently,
\emph{vision-language-action} (VLA) models have emerged as a broader
architectural class that maps multimodal inputs to robot
actions~\citep{ahn2022can, rt12022arxiv, driess2023palm,
rt22023arxiv, kim24openvla, black2024pi_0, kim2025fine,
intelligence2025pi05visionlanguageactionmodelopenworld, li2023vision,
figure2024helix, li2024cogact, liu2024rdt, wen2024tinyvla,
zhen20243dvla, bjorck2025gr00t, reuss2025flower, team2025gemini,
wen2025dexvla, zhou2025chatvlaunifiedmultimodalunderstanding}; most
VLAs are trained with IL-style objectives, though recent work also uses
RL fine-tuning~\citep{li2025vla, liu2026what}. A related direction is
the \emph{world action model} (WAM), which routes action prediction
through a learned dynamics
model~\citep{ye2026worldactionmodelszeroshot, hafner2019dream,
wu2023daydreamer, hafner2023mastering, hansen2024tdmpc2,
bruce2024genie, wu2024ivideogpt, hafner2025training,
gao2026dreamdojo, guo2026ctrlworld, bi2025motusunifiedlatentaction,
ye2026gigaworld}. These families differ in training and deployment assumptions, but
\textsc{Nautilus} targets their workflows rather than any single
model class.

\noindent\textbf{LLMs in robotics and embodied orchestration.}
Recent work has applied LLMs and VLMs inside robotics pipelines for
policy code generation, reward design, and embodied task
orchestration~\citep{liang2023code, ma2024eureka,
wang2023robogen, huang2023voxposer}. These systems place an LLM within
a policy, planning, or environment loop. \textsc{Nautilus} addresses a
different layer of the stack: not execution-time orchestration for a
single robot task, but research-time orchestration across
repositories, simulators, benchmarks, and robot interfaces.

\noindent\textbf{Benchmarks, infrastructure, and deployment stacks.}
Robot learning research is supported by a broad ecosystem of simulator
and policy frameworks~\citep{lerobot, isaaclab, robosuite,
Xiang_2020_SAPIEN, robomimic, jia2025x, mandlekar2023mimicgen, ros2,
Genesis, savva2019habitatplatformembodiedai}, benchmark
suites~\citep{liu2023libero, nasiriany2024robocasa, tao2024maniskill3,
james2019rlbench, chen2022towards, mees2022calvin, chen2025robotwin,
li2024behavior1k}, and large-scale
datasets~\citep{khazatsky2025droidlargescaleinthewildrobot, o2024open,
bu2025agibot_iros}. Recent integrations such as Isaac
Lab-Arena~\citep{isaaclab-arena2025} and evaluation suites such as
RoboArena~\citep{atreya2025roboarena} and
MolmoSpaces~\citep{kim2026molmospaces} connect parts of this stack and
support broader cross-policy comparison. Existing systems make
important contributions at the level of individual frameworks,
benchmark suites, or deployment pipelines, but are typically tied to
particular simulator families, benchmark families, or robot stacks.
\textsc{Nautilus} instead takes a contract-centric position,
standardizing interfaces between policies, benchmarks, and robots to
make cross-family composition and extension tractable without
rewriting framework internals. Table~\ref{tab:platforms} summarizes this comparison across coverage,
hardware interfaces, and extensibility.

%% file: sections/Tabel_compare_platforms.tex
\begin{table*}[t]
  \centering
  \caption{\textbf{Comparison of \sysname{} to representative
  robot learning platforms.} Rows are grouped into policy families,
  simulation benchmarks integrated under the typed
  Policy/Benchmark contracts, real-robot hardware interfaces (by
  embodiment class), and contracts demonstrably driven end-to-end by
  an LLM coding agent under our reproduction-validated protocol
  (\S\ref{sec:exp:reproduction}). Several peer platforms expose stable
  ABCs that an expert programmer can extend; in the bottom block we
  reserve \cmark{} for contracts that an LLM agent has executed
  end-to-end against a verified registry. \cmark{} indicates
  first-class support, \xmark{} no support, \pmark{} ad-hoc or
  community-contributed support without a stable interface.}
  \label{tab:platforms}
  \setlength{\tabcolsep}{5pt}
  \renewcommand{\arraystretch}{1.1}
  \resizebox{\textwidth}{!}{%
  \begin{tabular}{l|c|cccccc}
    \toprule
    \textbf{Capability} & \textbf{\sysname{}}
      & LeRobot~\citep{lerobot} & RoboTwin Platform~\citep{chen2025robotwin} & robomimic~\citep{robomimic} & robosuite~\citep{robosuite} & Isaac Lab~\citep{isaaclab} & Lab-Arena~\citep{isaaclab-arena2025} \\
    \midrule
    \multicolumn{8}{l}{\emph{Policy families}} \\
    \quad Imitation learning (IL)        & \cmark & \cmark & \cmark & \cmark & \xmark & \pmark & \cmark \\
    \quad Vision-Language-Action (VLA)   & \cmark & \cmark & \cmark & \xmark & \xmark & \xmark & \cmark \\
    \quad World Action Model (WAM)       & \cmark & \xmark & \xmark & \xmark & \xmark & \xmark & \xmark \\
    \quad Reinforcement learning (RL)    & \cmark & \cmark & \pmark & \xmark & \cmark & \cmark & \cmark \\
    \midrule
    \multicolumn{8}{l}{\emph{Benchmarks (simulation)}} \\
    \quad RoboCasa~\citep{nasiriany2024robocasa}                        & \cmark & \xmark & \xmark & \pmark & \xmark & \xmark & \xmark \\
    \quad LIBERO~\citep{liu2023libero}                          & \cmark & \cmark & \xmark & \xmark & \xmark & \xmark & \xmark \\
    \quad ManiSkill~\citep{tao2024maniskill3}                       & \cmark & \xmark & \xmark & \xmark & \xmark & \xmark & \xmark \\
    \quad RoboTwin~\citep{chen2025robotwin}                        & \cmark & \xmark & \cmark & \xmark & \xmark & \xmark & \xmark \\
    \quad ALOHA~\citep{zhao2023learning}                           & \cmark & \xmark & \xmark & \xmark & \xmark & \xmark & \xmark \\
    \midrule
    \multicolumn{8}{l}{\emph{Real-robot hardware interfaces}} \\
    \quad Single-arm                      & \cmark & \cmark & \cmark & \xmark & \xmark & \xmark & \xmark \\
    \quad Bimanual                        & \cmark & \cmark & \cmark & \xmark & \xmark & \xmark & \xmark \\
    \quad Dexterous hand                  & \cmark & \xmark & \xmark & \xmark & \xmark & \xmark & \xmark \\
    \quad Humanoid                        & \cmark & \cmark & \xmark & \xmark & \xmark & \pmark & \xmark \\
    \midrule
    \multicolumn{8}{l}{\emph{Automatically extendable to}} \\
    \quad New policies                    & \cmark & \xmark & \xmark & \xmark & \xmark & \xmark & \xmark \\
    \quad New benchmarks                  & \cmark & \xmark & \xmark & \xmark & \xmark & \xmark & \xmark \\
    \quad New robot interfaces            & \pmark & \xmark & \xmark & \xmark & \xmark & \xmark & \xmark \\
    \bottomrule
  \end{tabular}}
\end{table*}

%% file: sections/03_method.tex
\section{\sysname{}}
\label{sec:method}

\sysname{} is a Claude Code plugin that turns natural-language
robot learning requests into executed reproduction, evaluation,
finetuning, and deployment workflows. Its design has two layers. The
first is a \emph{substrate} of three engineering invariants---typed
interface contracts, chambered execution, and uniform inter-module
transport (\S\ref{sec:method:platform}). The second is a \emph{content
layer} of robotics-specific components---Guides, Sensors, State, and
workspace artefacts (\S\ref{sec:method:agent}). The corresponding
implementation stack is detailed in Appendix~\ref{app:method-impl:layers}
(Table~\ref{tab:layers}).

\subsection{Running example:
\texorpdfstring{$\pi_0$}{pi0} on LIBERO}
\label{sec:method:worked-example}
\begin{figure}[t]
\centering
\includegraphics[width=\linewidth]{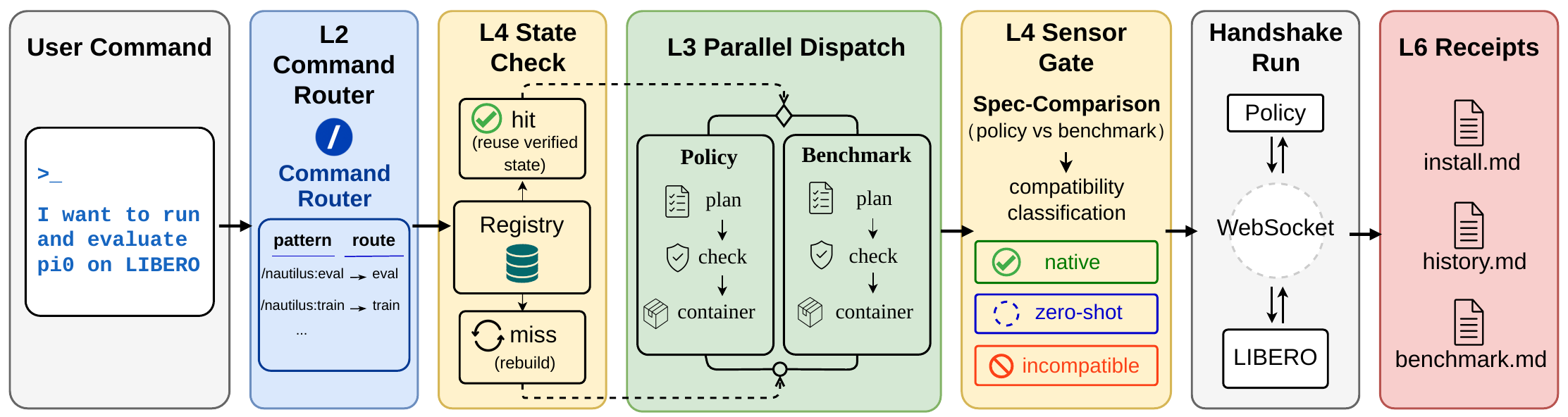}
\caption{Execution trace for the running example $\pi_0$ on
LIBERO (\S\ref{sec:method:worked-example}). The orchestrator routes a
natural-language request to \nautcmd{/nautilus:eval}; State, subagents,
Sensors, and workspace artefacts then turn the request into a
contract-checked, chambered, reproducible evaluation.}
\label{fig:method-trace}
\end{figure}

We use one running example throughout the section: a researcher says,
in natural language, ``I want to run and evaluate $\pi_0$ on the LIBERO
benchmark.'' Rather than asking the researcher to locate the right
repository entry points, align dependencies, and write the evaluation
glue, the orchestrator interprets the request and routes it to
\nautcmd{/nautilus:eval policy=pi0 benchmark=libero}. The resulting run
is the trace sketched in Figure~\ref{fig:method-trace}: the
orchestrator dispatches the task, State checks the registry, policy and
benchmark modules are generated in parallel, their contracts are
compared, a WebSocket smoke test is executed, and reproducibility
receipts are written.

\subsection{Substrate}
\label{sec:method:platform}




\sysname{} gives the agent a fixed substrate to build against. It has
three invariants: \emph{typed interface contracts}, which make policies, benchmarks, and
robots legible through fixed schemas; \emph{chambered execution}, which isolates generated artefacts
so that dependency failures do not leak across modules; and
\emph{uniform inter-module communication}, which gives simulated and real-robot
endpoints the same agent-visible observation/action transport.


\noindent\textbf{Typed interface contracts.}
\label{sec:method:interfaces}
A robot learning experiment is represented by three typed interfaces: a
policy, a robot embodiment, and a benchmark/environment. \emph{Policy}
exposes \texttt{infer(obs)\,$\to$\,actions} with an optional
\texttt{reset}. \emph{Robot} exposes \texttt{reset},
\texttt{get\_observation}, \texttt{apply\_action}, and
\texttt{safe\_stop}. \emph{Benchmark/Environment} exposes
\texttt{reset(seed)} and \texttt{step(action)} together with the task
definition, success criterion, and deterministic seeding used for the
published-reference checks in \S\ref{sec:exp:trust}. These contracts are
the unit of integration: adding a new policy, benchmark, or robot
requires satisfying the corresponding contract rather than modifying the
rest of the framework. In Figure~\ref{fig:method-trace}, the agent therefore connects $\pi_0$ and LIBERO through Policy and Benchmark/Environment contracts; for hardware experiments, a Robot contract fills the same role.


\noindent\textbf{Chambered execution.}
\label{sec:method:agent:sandbox}
Policy inference, simulator execution, and real-robot control often
depend on incompatible CUDA, PyTorch, MuJoCo/Isaac/SAPIEN, and ROS\,2
stacks. \sysname{} therefore runs generated artefacts in disposable
chambers: each chamber holds its own dependency environment while
exposing the same external contract. A \emph{policy container} serves
model inference, while a \emph{benchmark/environment container} hosts
the simulator or task suite; real hardware occupies the same
architectural slot from the agent's perspective. In the
$\pi_0$--LIBERO example, chambering allows policy and benchmark modules
to build in parallel without dependency failures leaking across them.


\noindent\textbf{Uniform inter-module communication.}
\label{sec:method:comm}
Contracts and chambers define what each module exposes, but modules
still need to exchange observations and actions. 
\sysname{} separates inter-module communication into two tiers.
\emph{Tier~1 (robot-local)} keeps high-bandwidth sensing and control in
the native robot stack (ROS\,2/shared memory). \emph{Tier~2
(off-robot)} is the only transport seen by the agent: a lightweight
WebSocket between a policy container and a benchmark or real-environment
endpoint. Simulators and robots therefore share one agent-visible
observation/action schema, while ROS\,2, shared memory, and other
robot-local communication mechanisms remain below the agent boundary.
In the $\pi_0$--LIBERO example, the WebSocket smoke test checks this
shared Tier~2 interface rather than a pair-specific communication path.


\subsection{Content layer}
\label{sec:method:agent}
On top of the substrate, the content layer supplies robotics-specific
procedures. We adopt the Guides+Sensors abstraction of harness
engineering~\citep{meng2026agentharness}: Guides steer work before
generation, while Sensors validate outputs afterward. \sysname{} adds
State to persist typed contracts across runs, and workspace artefacts to
record each run for reproducible hand-off. Operationally, \Lone{} is the
base agent loop, \Ltwo{} routes user intent, \Lthree{} executes
task-specific subagents, \Lfour{} validates artefacts and carries State,
\Lfive{} supplies domain references, and \Lsix{} records receipts. The
full six-layer implementation stack realising this content layer is given in
Table~\ref{tab:layers} (Appendix~\ref{app:method-impl:layers}).

\noindent\textbf{Guides.}
\label{sec:method:agent:skills}
Guides tell the agent how to act before it writes code. Instead of
inferring robot learning conventions from each repository, the agent is
given the relevant routing rules, subagent roles, schemas, and
validation protocols. They appear at three layers of
Table~\ref{tab:layers}: \Ltwo{} routes a request to a script or subagent
workflow; \Lthree{} runs subagents for policy servers, benchmark
wrappers, training scaffolds, and robot integrations; \Lfive{} provides
the shared reference library they consult (full inventory in
Appendix~\ref{app:skill-walkthroughs}).
In the worked example,
\nautcmd{/nautilus:eval policy=pi0 benchmark=libero} therefore means
``generate the policy and benchmark modules, then compare their registry
specs before launch.'' A \emph{spec} is the registry-side,
machine-readable form of the typed contract in
\S\ref{sec:method:interfaces}.

\noindent\textbf{Sensors.}
\label{sec:method:sensors}
Sensors validate what the agent generates. At the \Lfour{} boundary,
\sysname{} runs six checks---pre-action filtering, render-time auditing,
interface verification, spec comparison, tiered smoke testing, and
cross-run verification---each of which either triggers a local
correction or blocks the run; Appendix~\ref{app:sensors} gives the full
inventory.
In Figure~\ref{fig:method-trace}, the key Sensor is spec comparison. It
checks the $\pi_0$ Policy spec against the LIBERO Benchmark/Environment
spec before launch, surfacing observation-key or action-shape mismatches
before any heavyweight run. Registry promotion is handled separately by
the curation gate of \S\ref{sec:method:curation}.

\noindent\textbf{State.}
\label{sec:method:state}
State is the \Lfour{} persistent registry that lets \sysname{} reuse
verified module knowledge across runs.
Instead of rediscovering repository
layout, environment constraints, and policy/benchmark/robot interfaces
from scratch, the agent retrieves registry entries that bind pinned
code and environment artefacts to the typed specs of
\S\ref{sec:method:interfaces}. These specs describe the module boundary:
observation/action schemas, control mode, and benchmark criteria.
In the $\pi_0$--LIBERO trace, State is checked immediately after
routing. Verified entries support reuse and compatibility checks before
launch; unverified modules are generated against the same spec format
and can later enter curation. This turns policy--benchmark integration
from $\Theta(N\!\cdot\!M)$ hand-written adapters into $\Theta(N\!+\!M)$
machine-readable specs and a shared dispatch protocol.

\noindent\textbf{Workspace artefacts.}
\label{sec:method:workspace}
The \Lsix{} workspace layer records the evidence for a single run; it
is not itself a verified registry entry. \emph{Process logs} are
append-only notes in the target repository, while \emph{receipts} are
regenerated summaries at the repository root
(Appendix~\ref{app:method-impl:workspace}). Together, they record the
source commit, image ID, environment choices, benchmark protocol, key
decisions, and rerun recipe. For $\pi_0$ on LIBERO, the receipt makes
the run reproducible for a later user and provides evidence for a
possible registry submission.

\subsection{Registry curation and hardware support}
\label{sec:method:curation}
\label{sec:method:hardware}

The registry is a shareable index of reproducible entries, not a
centralized service. Each entry pins source and environment artefacts,
allowing any user to reproduce the same policy, benchmark, or robot
setup. The \texttt{verified} status is reserved for benchmark entries,
because benchmark protocols are far fewer in number and evolve more
slowly than policy configurations. We mark a benchmark entry as
\texttt{verified} only when it is supported by \emph{cross-policy
evidence}: several released policies, each run with its official
checkpoint through the same wrapper, reproduce their published
reference results with transparent deviations rather than a universal
pass/fail threshold.
\S\ref{sec:exp:trust} applies this criterion to the
initial set of verified entries; curation mechanics are detailed in
Appendices~\ref{app:method-impl:curation}
and~\ref{app:registry:curation}.

Hardware reuses the same substrate. \texttt{nautilus-collect} is a
chambered data-collection platform that implements the Robot typed
contract and exposes the same uniform transport. As a result, any
policy wrapped once by \sysname{} can be handed from a benchmark to a
real Robot endpoint for deployment. The platform also
covers robot data collection, training hand-off, and deployment. We
validate this path with a VLA manipulation rollout on a single-arm
Franka and an RL locomotion rollout on Unitree H1
(\S\ref{sec:exp:hardware}); details are in
Appendices~\ref{app:hardware} and~\ref{app:nautilus-collect}.

%% file: sections/04_experiments.tex
\section{Experiments}
\label{sec:experiments}

We organize experiments around three studies.
\S\ref{sec:exp:trust} evaluates benchmark-wrapper fidelity by
reproducing the reference results of public policies under their
released checkpoints. \S\ref{sec:exp:ablation} ablates the cost
of new policy onboarding on a controlled testbed,
isolating which components actually buy reusable
extension. \S\ref{sec:exp:hardware} closes the loop on real
hardware. The fine-tuning study and the communication-layer
microbenchmarks originally scoped alongside these studies are deferred
(see \S\ref{sec:discussion:future}).

\subsection{Benchmark trust validation}
\label{sec:exp:trust}

\noindent\textbf{RQ1.} Are our generated benchmark wrappers faithful to the references
they wrap?

\noindent\textbf{Protocol.} We instantiate the cross-policy evidence
protocol introduced in \S\ref{sec:method:curation}. The unit of
verification is a benchmark wrapper, not a single policy--benchmark
pair. For each candidate benchmark entry, we run several released
policies with their official checkpoints through the same generated
wrapper and compare the resulting scores against the corresponding
published references. The study covers imitation-learning,
vision-language-action, and world-action-model policies across LIBERO,
ManiSkill, RoboTwin, ALOHA, and RoboCasa. Because the reference sources
do not report a common across-seed variance, we do not impose a
universal pass/fail threshold. Instead, we report each reproduced score
and its deviation from the corresponding reference.
This study covers only the simulation row of
Table~\ref{tab:platforms}; real-robot deployment is evaluated
separately in \S\ref{sec:exp:hardware}.

\noindent\textbf{Metrics.}
Table~\ref{tab:trust_validation} reports, for each (method, benchmark)
pair, the wrapped success rate, the reference value from the original
publication, and the deviation $\Delta$ between them. No row is marked
with a fixed threshold-based pass/fail label.

\noindent\textbf{Findings.}
Across IL, VLA, and WAM policies, most wrapped results remain close to
their published references, indicating that the benchmark adapters
preserve the intended evaluation protocols. By construction, Table~\ref{tab:trust_validation}
proves cross-policy for the initial verified benchmark
entries: small deviations support reuse of the wrapper, while larger
deviations are retained transparently for future calibration.

\begin{table}[t]
  \centering
  \caption{\textbf{Cross-policy trust validation for benchmark
  wrappers.} Each row reports a (method, benchmark) pair: the
  success rate measured through our wrapper, the value reported in
  the original publication, and the deviation $\Delta$. Rows are
  grouped by policy family --- IL (\colorbox{ilrow}{\strut\,blue\,}),
  VLA (\colorbox{vlarow}{\strut\,green\,}), and WAM
  (\colorbox{wamrow}{\strut\,orange\,}).}
  \label{tab:trust_validation}
  \small
  \setlength{\tabcolsep}{3pt}
  \renewcommand{\arraystretch}{1.15}
  \setlength{\aboverulesep}{0pt}
  \setlength{\belowrulesep}{0pt}
  \begin{tabular}{@{}c p{0.25\linewidth} p{0.18\linewidth} c c c@{}}
    \toprule
    \textbf{Family} & \textbf{Method} & \textbf{Benchmark}
      & \textbf{Success rate} & \textbf{Reference} & \textbf{$\Delta$} \\
    \midrule
    \rowcolor{ilrow}
      &                                                                          & LIBERO~\citep{liu2023libero}      & $70.2$              & $72.4$              & $-2.2$ \\
    \rowcolor{ilrow}
      &                                                                          & ManiSkill~\citep{tao2024maniskill3} & $32.4$              & $30.2$              & $+2.2$ \\
    \rowcolor{ilrow}
      &                                                                          & RoboTwin~\citep{chen2025robotwin}  & $26.4$    & $28.0$   & $-1.6$ \\
    \rowcolor{ilrow}
      & \multirow{-4}{*}{Diffusion Policy~\citep{chi2025diffusion}}              & ALOHA~\citep{zhao2023learning}     & $75.8$              & $77.5$              & $-1.7$ \\
    \cmidrule(lr){2-6}
    \rowcolor{ilrow}
      &                                                                          & RoboTwin  & $30.0$    & $29.7$   & $+0.3$ \\
    \rowcolor{ilrow}
      \multirow{-6}{*}{IL} & \multirow{-2}{*}{ACT~\citep{zhao2023learning}}      & ALOHA     & $72.8$              & $72.3$              & $+0.5$ \\
    \midrule
    \rowcolor{vlarow}
      &                                                                          & LIBERO    & $93.6$              & $94.2$              & $-0.6$ \\
    \rowcolor{vlarow}
      & \multirow{-2}{*}{$\pi_0$~\citep{black2024pi_0}}                          & RoboTwin  & $42.6$  & $46.4$  & $-3.8$ \\
    \cmidrule(lr){2-6}
    \rowcolor{vlarow}
      &                                                                          & LIBERO    & $97.0$              & $96.8$             & $+0.2$ \\
    \rowcolor{vlarow}
      & \multirow{-2}{*}{$\pi_{0.5}$~\citep{intelligence2025pi05visionlanguageactionmodelopenworld}} & RoboCasa~\citep{nasiriany2024robocasa} & 18.6 & 16.9 & $+1.7$ \\
    \cmidrule(lr){2-6}
    \rowcolor{vlarow}
      &                                                                          & LIBERO    & $78.2$              & $76.5$              & $+1.7$ \\
    \rowcolor{vlarow}
      & \multirow{-2}{*}{OpenVLA~\citep{kim24openvla}}                           & ManiSkill & $4.0$              & $4.8$              & $-0.8$ \\
    \cmidrule(lr){2-6}
    \rowcolor{vlarow}
      \multirow{-7}{*}{VLA} & SmolVLA~\citep{shukor2025smolvla}                  & LIBERO    & $88.2$              & $87.3$              & $+0.9$ \\
    \midrule
    \rowcolor{wamrow}
      &                                                                          & LIBERO & $98.4$ & $98.5$ & $-0.1$ \\
    \rowcolor{wamrow}
      & \multirow{-2}{*}{cosmos-policy~\citep{kim2026cosmos}}                   & RoboCasa & $66.7$ & $67.1$ & $-0.4$ \\
    \cmidrule(lr){2-6}
    \rowcolor{wamrow}
      \multirow{-3}{*}{WAM} & Motus~\citep{bi2025motusunifiedlatentaction}     & RoboTwin & $86.9$ & $87.0$ & $-0.1$ \\
    \bottomrule
  \end{tabular}
\end{table}






\subsection{Ablation: from pairwise wiring to reusable extending}
\label{sec:exp:reproduction}
\label{sec:exp:ablation}

\paragraph{RQ2.} Can \sysname{} onboard new implementations and what is the cost?




\S\ref{sec:exp:trust} shows that, for existing policy --- benchmark implementation, our generated wrappers faithfully reproduce the reference performance. 
The harder question we care about more is whether the same level of trust can extend beyond existing integrations.

In practice, we find that a strong coding agent can often make a specific local setup work:
for example, wiring a policy implementation to a benchmark repository it does not yet support.
However, this success is usually local --- it is not sufficiently validated, deployable, or reusable across other benchmarks.
\sysname{} aims for a different goal.
Once an new component is onboarded through the shared interface,
it is not only connected to the current target but also automatically compatible with all the verified implementations and workflows already in the platform.


This section therefore studies whether such platform-aware onboarding can be achieved at reasonable cost while remaining faithful to reference performance.
We first introduce a controlled testbed that enables fair comparison between the raw coding-agent baseline and ablated \sysname{} variants.
Then we demonstrate with the components proposed in \S\ref{sec:method}, \sysname{} can achieve such extensibility at roughly the cost of a single one-off integration.

\paragraph{Testbed for extensibility}
\label{sec:exp:reproduction:testbed}
A direct validation on genuinely unseen policy--benchmark combination is difficult to audit:
without released checkpoints on the certain benchmark, success can only be judged after costly training.

We therefore construct a \emph{leave-one-benchmark-out} controlled proxy.
Starting from a public policy repository that already has a released checkpoint for a target benchmark,
we carefully remove the repository's existing support for that benchmark and ask the agent to reconstruct the missing integration.
This allows us to test a typical engineering task faced by robot learning researchers
--- connecting a custom policy repository to a not yet supported benchmark ---
while still retaining a task-level oracle:
the resulting evaluation should match the reference success rate within the trust band established in
\S\ref{sec:exp:trust}. Details are provided in
Appendix~\ref{app:method-impl:surgical-pruning}.

\label{sec:exp:ablation:axes}
We curate several pruned policy--benchmark pairs spanning IL and VLA policy families,
and covering benchmark suites already trust-validated in \S\ref{sec:exp:trust}.
 Detailed pairs are listed in Table~\ref{tab:ablation_per_pair}.
Each integration trial is performed in an isolated sandbox environment, where the agent initiates from a reset state with no global memory and can only reach the pruned source trees.
Table~\ref{tab:ablation_summary} reports the headline numbers
aggregated across the three pruned policy--benchmark pairs; the
per-pair breakdown for both models appears in
Table~\ref{tab:ablation_per_pair} in the appendix.

\paragraph{Metrics.}
Following \citet{han2026sweskillsbench} on evaluating coding-agent
performance, we measure and report:
\textbf{Wrap success} ($\uparrow$) measures whether the agent's
generated integration loads the released checkpoint and completes
the benchmark evaluation protocol end-to-end without manual
intervention.
\textbf{Task success} ($\uparrow$) further requires the resulting
score to lie within the trust band of \S\ref{sec:exp:trust}.
\textbf{Agent inference time} ($\downarrow$) measures cumulative
agent-side inference time spent on generation before the accepted
run.
\textbf{Interaction budget} ($\downarrow$) reports the total number
of agent turns required to reach a successful evaluation.
\textbf{Token/context budget} ($\downarrow$) reports total LLM
token usage and peak context length.
\textbf{Monetary cost} is the metered USD cost under the fixed
pricing table of each integration trial.

\begin{table*}[t]
  \centering
  \caption{\textbf{Comparison and ablations.}
  Four configurations are evaluated on the three pruned
  policy--benchmark pairs.
  Every cell reports paired values \textbf{Claude Opus 4.7 / Claude
  Sonnet 4.6}; the full per-pair breakdown appears in
  Table~\ref{tab:ablation_per_pair}. Time, interaction, token,
  context, and cost columns are averaged over accepted trials and
  across pairs with at least one accepted trial.}
  \label{tab:ablation_summary}
  \setlength{\tabcolsep}{4pt}
  \renewcommand{\arraystretch}{1.05}
  \resizebox{\textwidth}{!}{%
  \begin{tabular}{l cc c c cc c}
    \toprule
    &
    \multicolumn{2}{c}{\textbf{Agent success} ($\uparrow$)}
    & \textbf{Agent inference}
    & \textbf{Interaction budget} ($\downarrow$)
    & \multicolumn{2}{c}{\textbf{Token/context budget} ($\downarrow$)}
    & \textbf{Monetary cost} \\
    \cmidrule(lr){2-3}
    \cmidrule(lr){6-7}
    \textbf{Setting}
      & \textbf{Wrap} & \textbf{Task}
      & \textbf{Time}
      & \textbf{Agent turns}
      & \textbf{Tokens} & \textbf{Peak ctx}
      & \textbf{USD} \\
    \midrule

    \multicolumn{1}{l}{\textbf{Vanilla agent}}
      & 100.0\% / 98.8\% & 95.4\% / 93.2\%
      & 16m30s / 14m21s
      & 161.9 / 158.0
      & 12.2M / 12.4M & 123.1k / 130.0k
      & \$5.10 / \$5.19 \\

    \midrule
    \multicolumn{1}{l}{\textbf{\sysname{}}}
      & 100.0\% / 100.0\% & 98.7\% / 98.0\%
      & 23m51s / 17m00s
      & 152.4 / 115.5
      & 13.8M / 9.5M & 152.1k / 136.0k
      & \$6.36 / \$4.76 \\

    \midrule
    \multicolumn{8}{l}{\textbf{\sysname{} ablations}} \\
    w/o template
      & 55.3\% / -- & 55.1\% / --
      & 26m00s / --
      & 607.5 / --
      & 41.7M / -- & 219.2k / --
      & \$74.94 / -- \\

    w/o verified image
      & 100.0\% / 100.0\% & 87.6\% / 79.2\%
      & 25m18s / 15m40s
      & 431.2 / 287.5
      & 16.2M / 9.8M & 135.5k / 123.9k
      & \$14.80 / \$9.75 \\

    \bottomrule
  \end{tabular}}
\end{table*}

\paragraph{Findings.}
Table~\ref{tab:ablation_summary} suggests:
\textbf{Cost parity with the vanilla envelope.}
\sysname{} stays within the same cost envelope as the
vanilla agent, while delivering shared-interface
onboarding rather than a one-off $(P, B)$ wire-up. On
\texttt{Claude Opus 4.7}, tokens and peak context rise modestly
(13.8M vs.\ 12.2M; +30k peak ctx) because the skill bodies of
\S\ref{sec:method:agent:skills} are loaded before the first action,
but agent turns drop (152.4 vs.\ 161.9) suggesting the benefits of having pre-defined \S\ref{sec:method:interfaces}.
The longer inference time is mainly due to agent polling while waiting for validation tool calls under the Sensors stack
of \S\ref{sec:method:sensors};
On \texttt{Claude Sonnet 4.6} the comparison \sysname{} cost much less,
(115.5 vs.\ 158.0 turns, 9.5M vs.\ 12.4M tokens,
\$4.76 vs.\ \$5.19), evidence that the harness contribution grows
as the underlying model becomes weaker --- consistent with prior
observations in agent-harness research that scaffolding gains are
larger on weaker
backbones~\citep{lee2026metaharnessendtoendoptimizationmodel,
meng2026agentharness}.
\textbf{Without the typed interface template.}
The \texttt{w/o template} setting strips the Tier-2 policy
WebSocket schema of \S\ref{sec:method:comm}, collapsing the Opus
run to 55.3\% wrap success at 607.5 turns, 41.7M tokens, and
\$74.94, with the additional turns spent rediscovering the policy
WebSocket transport and hand-rolling pair-specific IPC plumbing ---
precisely the per-project glue work that the substrate of
\S\ref{sec:method:platform} is designed to remove.
\textbf{Without the verified registry.}
The \texttt{w/o verified image} setting forces \sysname{} to
regenerate the benchmark side instead of looking it up in the
State registry of \S\ref{sec:method:state}, roughly doubling the
workload, with Opus turns nearly tripling (152.4 to 431.2) and cost rising by
2.3$\times$; the configuration nonetheless passes a non-trivial
fraction of trials, indicating that the substrate alone supports
unaided dual-side wrapping.

\subsection{Real-robot deployment path}
\label{sec:exp:hardware}


\paragraph{Protocol and demonstration.}
This section evaluates the real-robot deployment path of \sysname{},
rather than reporting a new real-robot benchmark score. After a
checkpoint has been wrapped for benchmark evaluation, the same policy
artefact can be transferred to hardware by replacing the simulation
back-end with a hardware container, while keeping both the
policy-facing WebSocket endpoint and the platform-side container
unchanged. We demonstrate this path on two deliberately heterogeneous
systems: a VLA manipulation checkpoint deployed on a Franka arm and an
RL locomotion policy deployed on a Unitree H1. Appendix~\ref{app:hardware:deployment-details}
provides the hardware setup and the shared data-collection workflow.

For Franka, we take the reproduced $\pi_{0.5}$ wrapper from
Table~\ref{tab:trust_validation} and deploy it for a real
pick-and-place rollout without modifying the policy code. This
demonstrates that the same wrapper used in benchmark evaluation can be
reused in the hardware deployment path. The released checkpoint,
however, does not zero-shot solve our real setup. When task-level
adaptation is required, \texttt{nautilus-collect} supports the same
collect--fine-tune--redeploy workflow under the Robot contract.
\label{sec:exp:hardware:franka}

For Unitree H1, we deploy an RL-trained locomotion policy through the
same policy-facing endpoint and platform-side container. Compared with
the benchmark setup, only the H1 hardware container and launch namespace
are changed.



\begin{figure}[!t]
  \centering
  \begin{subfigure}[t]{0.4\linewidth}
    \centering
    \includegraphics[width=\linewidth]{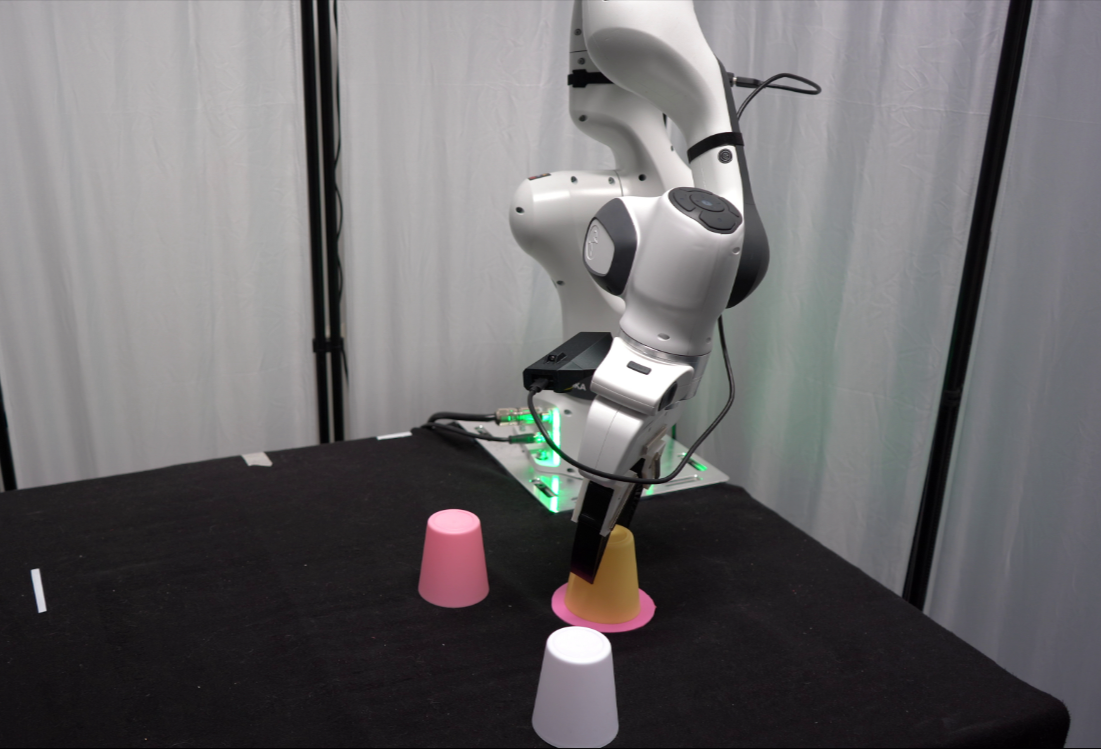}
    \caption{$\pi_{0.5}$ wrapper deployed on a Franka Panda arm.}
    \label{fig:hardware:franka}
  \end{subfigure}
  \hspace{0.02\linewidth}
  \begin{subfigure}[t]{0.4\linewidth}
    \centering
    \includegraphics[width=\linewidth]{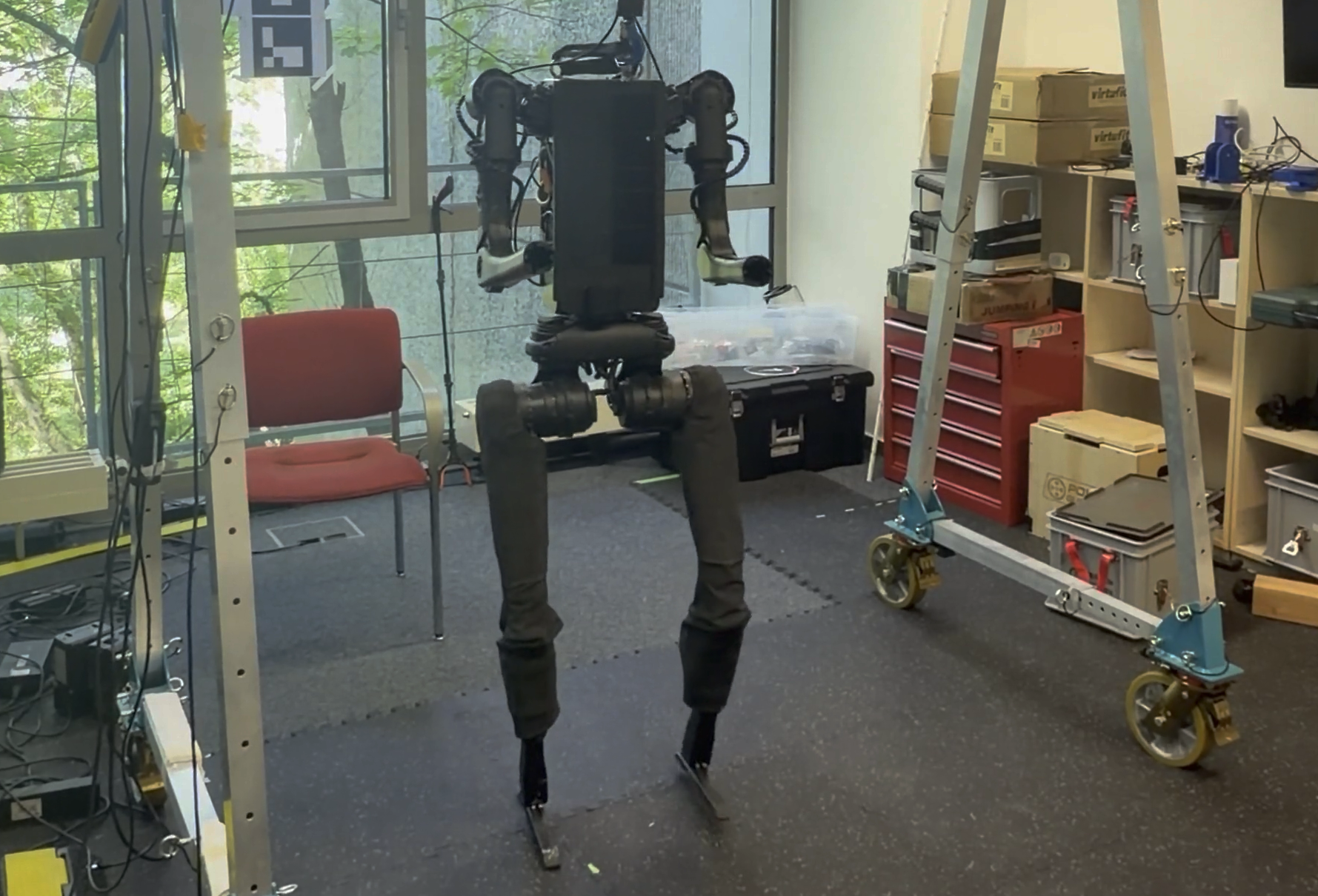}
    \caption{RL-trained locomotion policy deployed on Unitree H1.}
    \label{fig:hardware:h1}
  \end{subfigure}
  \caption{\textbf{Benchmark-to-robot deployment under a shared
  substrate.} A policy checkpoint wrapped for benchmark evaluation can
  be reused on hardware through the same policy-facing WebSocket
  endpoint and platform-side \texttt{Robot} contract. The deployment
  changes only the hardware container and launch namespace:
  \textbf{(\subref{fig:hardware:franka})} a reproduced $\pi_{0.5}$
  wrapper on a Franka Panda arm, and
  \textbf{(\subref{fig:hardware:h1})} an RL-trained locomotion policy
  on Unitree H1.}
  \label{fig:hardware}
\end{figure}

\paragraph{Takeaway.}
\label{sec:exp:hardware:synthesis}
These demonstrations test substrate portability, not cross-policy
transfer or state-of-the-art robot performance. The point is that any
checkpoint wrapped behind the shared \texttt{Policy} interface can be
iterated from benchmark evaluation to real hardware without a separate
policy-integration effort. Policy-family and embodiment-specific
details remain confined below the \texttt{Robot} contract, while the
agent-facing and policy-facing surfaces remain fixed.

%% file: sections/05_discussion.tex
\section{Discussion and Future Work}
\label{sec:discussion}

\textbf{Discussion.}
Our results reinforce the central claim of \S\ref{sec:intro}: the
current constraint on agentic robot learning automation is the harness,
not the agent model. The ablations in
\S\ref{sec:exp:ablation:axes} show that removing any single content
component---Guides, Sensors, or State---causes a much larger drop in
end-to-end success than swapping among LLMs. The real-robot
deployment demonstrations in \S\ref{sec:exp:hardware} extend the same
conclusion across the sim-to-real boundary: the shared Policy and Robot
contracts let a benchmark-wrapped policy move to heterogeneous hardware
without a separate integration path.

The current system also has clear limits. First, coverage is limited to
the workflow classes we implemented, namely reproduction, evaluation,
deployment, and initial finetuning support; broader research workflows
still depend on extending the skills library. Second, our cross-simulator
unification is at the environment-API level rather than the physics
layer, so simulator-specific effects should still be reported per
simulator. Third, because \textsc{Nautilus} is
implemented as a plugin on top of a commodity coding-agent harness, it
inherits some upstream platform churn, which we mitigate but do not
eliminate.

\textbf{Future work.}
\label{sec:discussion:future}
The most important next step is an auto-research loop for robot
learning. Today \textsc{Nautilus} automates the engineering loop around
a given $(P,B,R)$ triple; the natural extension is to let the agent
propose experiments, run ablations, interpret results, and refine
hypotheses using the same substrate. A practical step in this direction
is to close the train--eval--deploy loop more fully, including
automated hyperparameter selection, multi-seed sweeps, and
reproduction-grade report generation. We also see automatic skill
induction from agent--human traces as a promising way to expand
coverage beyond hand-authored procedures, and broader hardware support
plus human-in-the-loop safeguards as important system-level extensions.
More broadly, the substrate$+$(Guides, Sensors, State) anatomy may
transfer to adjacent scientific domains where generic coding harnesses
still fail, though validating that claim requires a second
domain-specific instantiation.

%% file: sections/06_conclusion.tex
\section{Conclusion}
\label{sec:conclusion}

We argued that the current constraint on agentic automation of robot
learning research is structural: a code-writing LLM cannot orchestrate
research workflows at scale when the surrounding hardware/software stack offers
no stable surface to target. \sysname{} responds with a co-designed
substrate---typed contracts, chambered execution, and uniform
inter-module transport---together with a content layer of Guides,
Sensors, State, and reproducibility artefacts. Across benchmark
verification, reusable onboarding at the cost of roughtly an one-off integration, 
and real-robot deployment demonstrations, the system turns fragmented research workflows into
structured agentic procedures rather than pair-specific glue code. We
will release the platform, skills library, benchmark suite, and
cross-policy evidence protocol as a community artefact. More broadly,
\sysname{} suggests a form of progress in which expert-built harnesses
do not simply make models replace experts; they make domain expertise
executable, reusable, and inspectable. 
How the relationship between experts and models should evolve remains an open question, but our work takes one small step in that co-evolutionary direction: experts encode the structure of a field into the harness, while the harness scales that structure across workflows and boosts back to experts' workflows. 



%% file: sections/A_appendix.tex
\section{Module interfaces and the policy WebSocket}
\label{app:interfaces}

This appendix supplements \S\ref{sec:method:interfaces} and
\S\ref{sec:method:comm}: the three typed interfaces are engineered so
that any policy can communicate with any benchmark or real-robot
endpoint through a \emph{single} Tier-2 WebSocket protocol.
Algorithm~\ref{alg:endpoints} shows the two minimal
endpoints---policy-side on the left, environment-side on the
right---and the subsection that follows documents the
\texttt{policy\_websocket} package they both depend on. The two
endpoints are condensed from the reference implementations released
alongside the platform
(\texttt{openvla/vla-scripts/policy\_server.py} and
\texttt{LIBERO/scripts/run\_eval.py}); substituting either side leaves
the other unchanged.

\subsection{Typed contract overview}
\label{app:interfaces:contracts}

The Method section uses \emph{contract} to mean the small typed surface
that an agent must satisfy before a module can enter the registry. Each
contract hides implementation-specific machinery and exposes only the
methods needed for composition. Algorithm~\ref{alg:typed-contracts}
summarizes the three surfaces used throughout the paper; the later
appendix subsections expand them into the reference WebSocket,
benchmark, and hardware implementations.

\begin{algorithm}[t]
\centering
\caption{Three typed contracts used as the module boundary. Policy
implementations hide model-specific loading and preprocessing;
Benchmark/Environment implementations hide simulator-specific task
logic and success criteria; Robot implementations hide drivers,
controllers, and safety logic.}
\label{alg:typed-contracts}
\setlength{\tabcolsep}{0pt}
\renewcommand{\arraystretch}{1.08}
\begin{tabular}{@{}p{0.31\linewidth}@{\hspace{0.9em}}p{0.31\linewidth}@{\hspace{0.9em}}p{0.31\linewidth}@{}}
\begin{minipage}[t]{\linewidth}
\begin{lstlisting}[style=nautilus]
# === Policy ===
class Policy:
    def reset(self) -> None:
        ...

    def infer(
        self,
        obs: Observation,
    ) -> Action:
        ...
\end{lstlisting}
\end{minipage}
&
\begin{minipage}[t]{\linewidth}
\begin{lstlisting}[style=nautilus]
# === Benchmark / Env ===
class Benchmark:
    def reset(
        self, seed: int,
    ) -> Observation:
        ...

    def step(
        self, action: Action,
    ) -> Transition:
        ...
\end{lstlisting}
\end{minipage}
&
\begin{minipage}[t]{\linewidth}
\begin{lstlisting}[style=nautilus]
# === Robot ===
class Robot:
    def reset(
        self,
    ) -> Observation:
        ...

    def get_observation(
        self,
    ) -> Observation:
        ...

    def apply_action(
        self, action: Action,
    ) -> None:
        ...

    def safe_stop(self) -> None:
        ...
\end{lstlisting}
\end{minipage}
\\
\end{tabular}
\end{algorithm}

The registry stores what the agent needs to compose these modules:
observation keys, action shape, timing, reset semantics, and
verification status. Algorithm~\ref{alg:endpoints} and
Table~\ref{tab:interfaces:protocol} give the Policy WebSocket
implementation; Algorithm~\ref{alg:benchmark-contract} gives the
Benchmark/Environment surface; and Algorithm~\ref{alg:robot-contract}
gives the hardware-facing Robot surface.

\begin{algorithm}[t]
\centering
\caption{Two minimal endpoints of the policy WebSocket protocol. The
policy server (left) wraps any inference-capable model behind
\texttt{BasePolicy} (optionally through \texttt{ActionChunkBroker}
for chunk-prediction models). The policy client (right) is itself a
\texttt{BasePolicy}, so a benchmark or real-robot rollout loop
drives a local stub and a remote checkpoint with identical code.
Substituting either side leaves the other unchanged.}
\label{alg:endpoints}
\setlength{\tabcolsep}{6pt}
\renewcommand{\arraystretch}{1.15}
\begin{tabular}{@{}p{0.475\linewidth}@{\hspace{1.5em}}p{0.475\linewidth}@{}}
\begin{minipage}[t]{\linewidth}
\begin{lstlisting}[style=nautilus]
# === Policy server ===
from policy_websocket import (
    BasePolicy, WebsocketPolicyServer,
    ActionChunkBroker,
)

class MyPolicy(BasePolicy):    # wrap any model
    def __init__(self, ckpt):
        self.model = load_model(ckpt)
    def infer(self, obs):      # core contract
        x = preprocess(obs)
        a = self.model.predict(x)  # (H, act_dim)
        return {"actions": a}

# optional: chunk policy -> per-step server
policy = ActionChunkBroker(
    MyPolicy(ckpt), action_horizon=H,
)
WebsocketPolicyServer(         # serve over ws://
    policy=policy,
    host="0.0.0.0", port=PORT,
    metadata={"action_dim": D,
              "execute_steps": H},
).serve_forever()
\end{lstlisting}
\end{minipage}
&
\begin{minipage}[t]{\linewidth}
\begin{lstlisting}[style=nautilus]
# === Policy client (benchmark / robot) ===
from policy_websocket import (
    WebsocketClientPolicy,
)

# remote policy behaves like a local one
policy = WebsocketClientPolicy(
    host=HOST, port=PORT,
)
meta = policy.get_server_metadata()
assert meta["action_dim"] == D

env = make_env()      # sim env or real robot
for ep in range(N_episodes):
    obs = env.reset(); policy.reset()
    for t in range(T_max):
        act = policy.infer(obs)["actions"]
        obs, _, done, _ = env.step(act)
        if done: break
\end{lstlisting}
\end{minipage}
\\
\end{tabular}
\end{algorithm}

\subsection{The \texttt{policy\_websocket} package}
\label{app:interfaces:protocol}

\sysname{} ships the Tier-2 protocol of \S\ref{sec:method:comm} as a
stand-alone library, \texttt{policy\_websocket}, with no dependency
on the rest of the platform. The protocol surface is exhausted by
the four classes in Table~\ref{tab:interfaces:protocol} and the
session state machine described below.

\begin{table}[t]
\centering
\caption{The four classes that make up the
\texttt{policy\_websocket} public API. Together they specify the
entire surface that the two endpoints in
Algorithm~\ref{alg:endpoints} depend on.}
\label{tab:interfaces:protocol}
\setlength{\tabcolsep}{6pt}
\renewcommand{\arraystretch}{1.15}
\begin{tabular}{@{}p{0.22\linewidth} p{0.34\linewidth} p{0.36\linewidth}@{}}
\toprule
\textbf{Class} & \textbf{Role} &
\textbf{Key methods / behaviour} \\
\midrule
\texttt{BasePolicy} &
abstract contract every policy implements &
\texttt{infer(obs)}\,$\to$\,\texttt{\{"actions":\,...\}}; optional
\texttt{reset()} \\
\texttt{Websocket\-PolicyServer} &
wraps any \texttt{BasePolicy} and serves it over WebSocket &
\texttt{serve\_forever()}; pushes a \texttt{metadata} blob on
connect; injects \texttt{server\_timing} into every response \\
\texttt{Websocket\-ClientPolicy} &
itself a \texttt{BasePolicy}; forwards \texttt{infer} to a remote
server &
\texttt{infer(obs)},
\texttt{get\_server\_metadata()},
\texttt{close()} --- drop-in substitute for an in-process policy \\
\texttt{Action\-ChunkBroker} &
adapts \emph{predict-$N$, execute-$M$} chunked policies to one-step
clients &
wraps any \texttt{BasePolicy}; queries the inner policy once per
$M$ \texttt{infer} calls (ACT, Diffusion Policy, OpenVLA-OFT) \\
\bottomrule
\end{tabular}
\end{table}

\paragraph{Session state machine.} A session has three phases.
\emph{(i) Connect}: the client opens a WebSocket; the server
immediately pushes its \texttt{metadata} blob (\texttt{action\_dim},
\texttt{policy\_name}, \texttt{checkpoint}, \ldots) so the client
can validate compatibility before the first step.
\emph{(ii) Loop}: at each step the client sends one msgpack-encoded
observation dict and receives one action dict; the response carries
a \texttt{server\_timing} field (\texttt{infer\_ms},
\texttt{prev\_total\_ms}) so the client can profile inference
latency without instrumenting the server.
\emph{(iii) Termination}: the client closes the connection at
episode end. If \texttt{infer} raises on the server side, the server
sends the formatted traceback as a string and closes the connection
with code \texttt{INTERNAL\_ERROR}; the client surfaces this as a
\texttt{RuntimeError}, so policy-side failures propagate to the
rollout loop instead of being silently swallowed.

\paragraph{Wire format.} Payloads are serialised with msgpack plus a
NumPy extension, which carries multi-camera RGB-D observations and
action arrays without Base64 inflation; the connection is persistent
and on localhost the round-trip is approximately $0.3\,\text{ms}$.
A \texttt{GET /healthz} endpoint on the same port serves liveness
probes for container orchestration.

\paragraph{What the symmetry buys.} Both endpoints in
Algorithm~\ref{alg:endpoints} depend only on
\texttt{policy\_websocket}; neither imports a symbol from the other.
The consequence is operational: a new policy is added by writing the
left column without touching any benchmark or robot code, and a new
benchmark or real-robot stack is added by writing the right column
without touching any policy. Any (policy, benchmark) or (policy,
robot) pair therefore composes without a custom adapter --- a
random-action stub on the left can drive any client for L3-IL smoke
testing (\S\ref{app:benchmark-skill:smoke}), and the same simulator
client can drive any registered checkpoint on the right. This is
also what permits \sysname{}'s \texttt{policy-generator} and
\texttt{benchmark-generator} skills (\S\ref{sec:method:agent:skills})
to scaffold the two sides independently: they target the same wire
format, share no state, and the wire format is the entirety of the
contract.

\paragraph{Deployed instances.} Concretely, the policy server in
Algorithm~\ref{alg:endpoints} has been exercised in tree with
OpenVLA-OFT, OpenPI-$\pi_0$, and Diffusion Policy checkpoints; the
policy client with the LIBERO, RoboCasa, and ManiSkill simulators and
with real-robot rollouts driven through \texttt{nautilus-collect}
(\S\ref{app:nautilus-collect:agentboundary}). None of these combinations
required modifications to either column of
Algorithm~\ref{alg:endpoints}; the per-checkpoint specifics
(observation remapping, action chunk size, gripper sign convention)
live entirely \emph{inside} the policy-server class, behind the
\texttt{infer} method.

\subsection{Benchmark/Environment contract}
\label{app:interfaces:benchmark}

The benchmark side of every Tier-2 WebSocket session is a
Gym-compatible environment with an associated task definition and
success criterion (Algorithm~\ref{alg:benchmark-contract}). Any
benchmark wrapper that satisfies this two-method surface --- LIBERO,
RoboCasa, ManiSkill, RoboTwin, and ALOHA in this submission --- is
mountable through
\texttt{policy\_websocket}'s \texttt{WebsocketClientPolicy} (right
column of Algorithm~\ref{alg:endpoints}) without further modification.
The success criterion is benchmark metadata used to aggregate rollouts
on the standard task split, the same surface that \S\ref{sec:exp:trust}
validates against published references.

\begin{algorithm}[t]
\caption{The two-method Benchmark/Environment contract every
supported task suite satisfies above its simulator backend (Isaac~Sim,
MuJoCo, or SAPIEN). \texttt{reset} and \texttt{step} mirror the Gym
API; task definition, success criterion, and deterministic seeding are
stored as benchmark metadata.}
\label{alg:benchmark-contract}
\begin{lstlisting}[style=nautilus]
class Benchmark:
    def reset(self, seed: int) -> Observation: ...
    def step(self, action: Action) -> Transition: ...
\end{lstlisting}
\end{algorithm}

\section{Robot hardware platform}
\label{app:hardware}

\paragraph{Design principle.} Hardware is where the modularity claim of
\S\ref{sec:method:platform} is most aggressively stress-tested: each
robot family carries its own driver ecosystem, control-loop frequency,
joint limits, and safety logic, and these concerns do not generalise
across embodiments. Where DROID~\citep{khazatsky2025droidlargescaleinthewildrobot} standardises
the \emph{box}---one Franka, one camera rig, one teleoperation
stack---so that data-collection sites can replicate the same setup,
\sysname{} takes the deliberate inverse position: we standardise the
\emph{interface} and let the same agent stack drive heterogeneous
embodiments without modifying policy or benchmark code. Replicability
of any single embodiment is delegated to whoever ships its driver;
portability across embodiments is delegated to the contract.

\paragraph{Two-container isolation.} \sysname{} factors every supported
platform into two ROS\,2-based containers whose only coupling is a fixed
upward message schema. The \emph{hardware container} hosts whatever talks
to the manufacturer driver and re-publishes everything upward as the same
standardised ROS\,2 hardware interface---a fixed set of state topics,
command topics, services, and actions whose schema does not depend on the
robot underneath. The container is deliberately permissive about
\emph{how} that driver is implemented: a native ROS\,2 package, a Python
SDK wrapper around a vendor library, or a C++ binary all sit equally well
inside, provided whatever sits inside speaks the standard schema upward.
The \emph{platform container} hosts the robot learning stack---perception,
policy inference, and the Tier-2 WebSocket endpoint of
\S\ref{sec:method:comm}---and consumes only that uniform interface; it
never imports a manufacturer symbol or links against a vendor library.
The two communicate over DDS plus host-shared-memory IPC, which insulates
the deep-learning stack's dependency closure from the driver's---these
are otherwise incompatible at the system-library level---and confines
cross-version friction to a one-time build-time problem. ROS\,2 Humble on
Ubuntu~22.04 is the standard base for both containers; the only exception
we currently ship is Franka with Polymetis, where the hardware container
is pinned to Ubuntu~20.04 / ROS\,2 Foxy purely for Polymetis
compatibility---the platform container above it stays on the standard
base. Three loop frequencies coexist on this fabric: the manufacturer
driver runs at 1\,kHz, ROS\,2 state aggregation at 50\,Hz, and the
policy/benchmark application loop at 15\,Hz, which is sufficient headroom
for diffusion- and VLA-class policies whose inference dominates the
application-loop budget.

\paragraph{The \texttt{Robot} contract.} Above this uniform hardware
interface, the platform-container side of every supported robot exposes
the same four-method \texttt{Robot} contract introduced in
\S\ref{sec:method:interfaces} (Algorithm~\ref{alg:robot-contract}).

\begin{algorithm}[t]
\caption{The four-method \texttt{Robot} contract that every supported
embodiment implements above the standardised ROS\,2 hardware interface.}
\label{alg:robot-contract}
\begin{lstlisting}[style=nautilus]
class Robot:
    def reset(self) -> Observation: ...
    def get_observation(self) -> Observation: ...
    def apply_action(self, action: Action) -> None: ...
    def safe_stop(self) -> None: ...
\end{lstlisting}
\end{algorithm}

\noindent Each method is a thin reader or writer of the standardised
hardware interface: \texttt{get\_observation} consumes the standardised
state topics, \texttt{apply\_action} publishes the standardised command
topics or invokes the standardised services, \texttt{reset} composes them
into a known initial state, and \texttt{safe\_stop} triggers the
standardised emergency-stop service. Everything below the contract---%
manufacturer driver, real-time inverse kinematics, joint-limit
enforcement, fault recovery---is reused from the existing ROS\,2 driver
ecosystem rather than reimplemented. This deliberate non-invention is
what keeps the integration cost of a new embodiment bounded, and it is
what allows the agent's robot-skill family
(\S\ref{sec:method:agent:skills}) to target a single schema regardless of
which robot is on the other end.

\paragraph{Currently supported platforms.} The platform integrates
four robot families that together span the four embodiment classes
introduced in \S\ref{sec:method:interfaces}; the same \texttt{Robot}
contract serves all four
(Table~\ref{tab:hardware:platforms}).

\begin{table}[t]
\centering
\caption{Robot families currently integrated under the four-method
\texttt{Robot} contract of \S\ref{sec:method:interfaces}. The driver
column illustrates the deliberate heterogeneity allowed inside the
hardware container (ROS\,2 package, Python SDK, or C++ SDK); the
upward ROS\,2 schema is identical regardless. From the agent's
WebSocket-side interface (\S\ref{sec:method:comm}), a single-arm xArm
rollout and a humanoid H1 rollout are indistinguishable apart from
their observation and action shapes.}
\label{tab:hardware:platforms}
\setlength{\tabcolsep}{6pt}
\renewcommand{\arraystretch}{1.15}
\begin{tabular}{@{}p{0.18\linewidth} p{0.20\linewidth} p{0.24\linewidth} p{0.28\linewidth}@{}}
\toprule
\textbf{Platform} & \textbf{Embodiment class} &
\textbf{Driver inside hardware container} & \textbf{Container base} \\
\midrule
Franka Panda 7-DoF &
single-arm and bimanual manipulation &
Polymetis~\citep{lin2021polymetis} (Python over libfranka), 1\,kHz &
Ubuntu 20.04 / ROS\,2 Foxy --- pinned for Polymetis compatibility \\
UFactory xArm 7 &
single-arm manipulation &
xArm Python SDK &
Ubuntu 22.04 / ROS\,2 Humble \\
Allegro Hand (16-DoF) &
dexterous manipulation (mounted on a host arm) &
Allegro ROS\,2 driver (native ROS\,2 package) &
Ubuntu 22.04 / ROS\,2 Humble \\
Unitree H1 &
humanoid locomotion &
Unitree SDK (C++) &
Ubuntu 22.04 / ROS\,2 Humble \\
\bottomrule
\end{tabular}
\end{table}

\paragraph{Adding a new embodiment.} Bringing a new robot under
\sysname{} requires four artefacts: a \texttt{Robot} subclass
implementing the four contract methods, a URDF (or equivalent
kinematic description), a ROS\,2 node that wraps the manufacturer
driver and exchanges the standard message schema, and a launch file
that wires the two containers together. The robot-skill family of
\S\ref{sec:method:agent:skills} is best understood as
\emph{scaffolding} for this process rather than full automation: it
ships a triggerable code base in which the four artefacts appear as
templates with the contract methods, ROS\,2 schema, and launch
wiring already laid out. The embodiment-specific judgement---which
SDK call corresponds to which contract method, what the joint
limits and safe default impedance gains should be, how the
manufacturer driver reports faults---still has to be supplied by a
human (or by a human-in-the-loop agent run) on a per-platform basis.
With the scaffolding in place we have found the integration scope to
fit within a few hundred lines per platform, configuration included;
the structural simplicity of the target is what bounds that scope,
not the skill family producing finished code on its own.

\section{Robot learning platform: \texttt{nautilus-collect} implementation
and data-collection stack}
\label{app:nautilus-collect}

This appendix is the concrete realisation of the on-robot Tier-1
transport of \S\ref{sec:method:comm} and of the typed \texttt{Robot}
contract of \S\ref{sec:method:interfaces}. Where \S\ref{app:hardware}
described the hardware-side platform list, the four-method contract,
and the cost of onboarding a new embodiment, this appendix describes
what runs \emph{inside} those containers and the data-collection stack
that ships on top. The argument back to \S\ref{sec:method:platform}
is that \texttt{nautilus-collect} is empirical evidence of modularity at
the right grain --- and, as a direct consequence, that the same code
path serves both teleoperated data collection and closed-loop policy
rollout, with only the action source swapped.

\subsection{Modular node architecture}
\label{app:nautilus-collect:nodes}

The runtime decomposes into five ROS\,2 nodes, each owning exactly
one device --- arm, gripper, state aggregator, teleoperation reader,
and hand-eye TF publisher. Each node prefixes its topics and services
with a configurable namespace, which is the mechanism behind the
bimanual variant referenced in \S\ref{app:hardware}: a bimanual rig
is two single-arm node groups running under different namespaces with
no code duplication, and the aggregator is configured by namespace
list rather than by arm count. From the agent's perspective the
namespace prefix is the only coupling between nodes.

\subsection{ROS\,2 interface schema}
\label{app:nautilus-collect:schema}

The wire-level instantiation of the typed contract of
\S\ref{sec:method:interfaces} comprises four entry families: a
canonical \texttt{RobotState} message that, after the WebSocket layer
of \S\ref{sec:method:comm} strips the ROS\,2 envelope, becomes
exactly the \texttt{Observation} the agent sees; control-mode
services that switch impedance modes in one call rather than as a
code edit; kinematics services that expose IK and FK so
agent-generated code never has to ship a solver of its own; and a
long-running action with progress feedback, demonstrating the
action-vs-service split. The schema is deliberately symmetric across
embodiments --- the bimanual driver, and the xArm, Allegro Hand, and
Unitree H1 wrappers all speak it --- which is why a single agent
skill family can target every embodiment without per-platform
branching.

\subsection{Python module layer}
\label{app:nautilus-collect:python}

The Python layer is the \emph{narrow waist} of the platform:
everything below it is ROS\,2, everything above it is policy or
agent code. It declares the four-method \texttt{Robot} contract of
\S\ref{sec:method:interfaces} (reproduced in \S\ref{app:hardware})
above the schema below, ships two Franka implementations in
tree (single-arm and bimanual, each of order a few hundred lines,
consistent with the integration scope claimed in
\S\ref{app:hardware}), and exposes a Gym-compatible adapter that
lets the same \texttt{Robot} face either a human teleoperator or a
policy server. The only difference between the two clients is who
calls \texttt{apply\_action}, which is the structural reason that
the data-collection stack of \S\ref{app:nautilus-collect:datacollect} and
the closed-loop rollout share their entire infrastructure below the
action source.

\subsection{Data-collection stack}
\label{app:nautilus-collect:datacollect}

Data collection is the most exercised loop in \texttt{nautilus-collect}:
the same stack has been used to gather manipulation trajectories on
the Franka single-arm and bimanual lines that constitute the
reference implementation in tree, and is reused at inference time
with only the action source swapped from the human teleoperator to a
policy server. We document the stack here because, in addition to
being a useful artefact in its own right, it is the most operationally
exercised piece of evidence for the modularity claim of
\S\ref{sec:method:platform}.

\begin{figure}[t]
  \centering
  \begin{subfigure}[t]{0.48\linewidth}
    \centering
    \includegraphics[width=\linewidth]{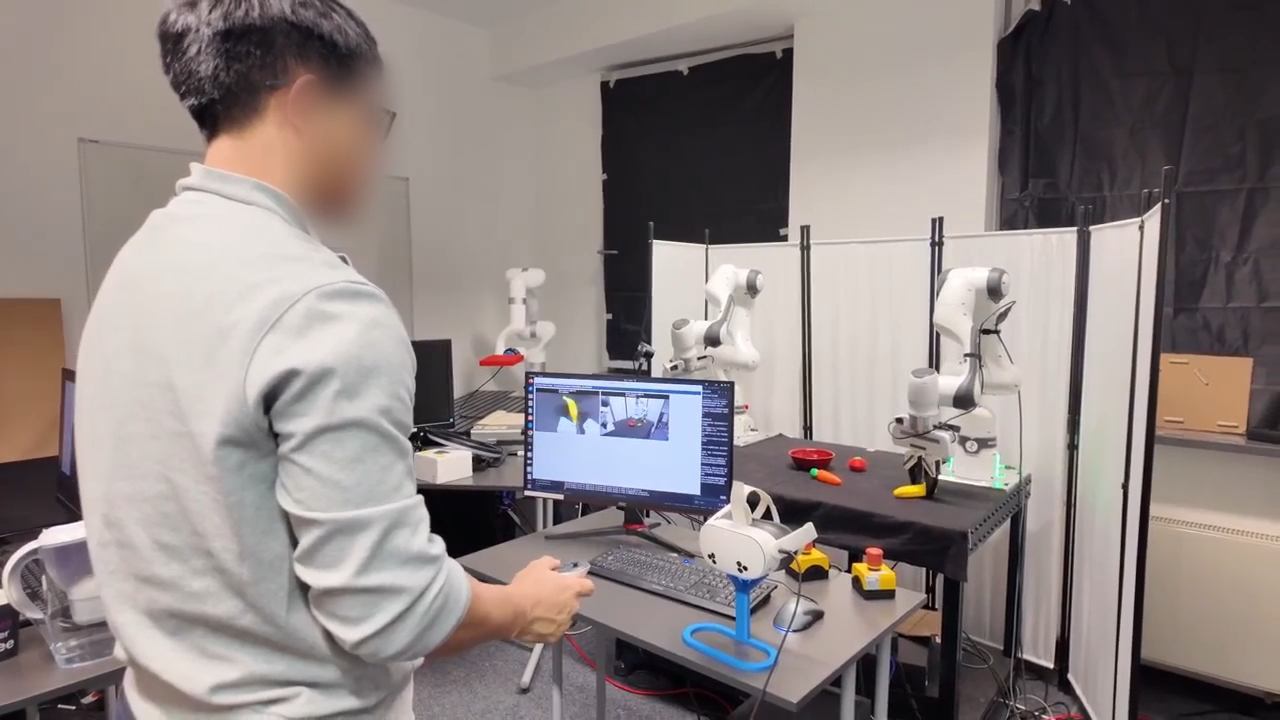}
    \caption{VR teleoperation via Oculus Quest controllers.}
    \label{fig:teleop:vr}
  \end{subfigure}
  \hfill
  \begin{subfigure}[t]{0.48\linewidth}
    \centering
    \includegraphics[width=\linewidth]{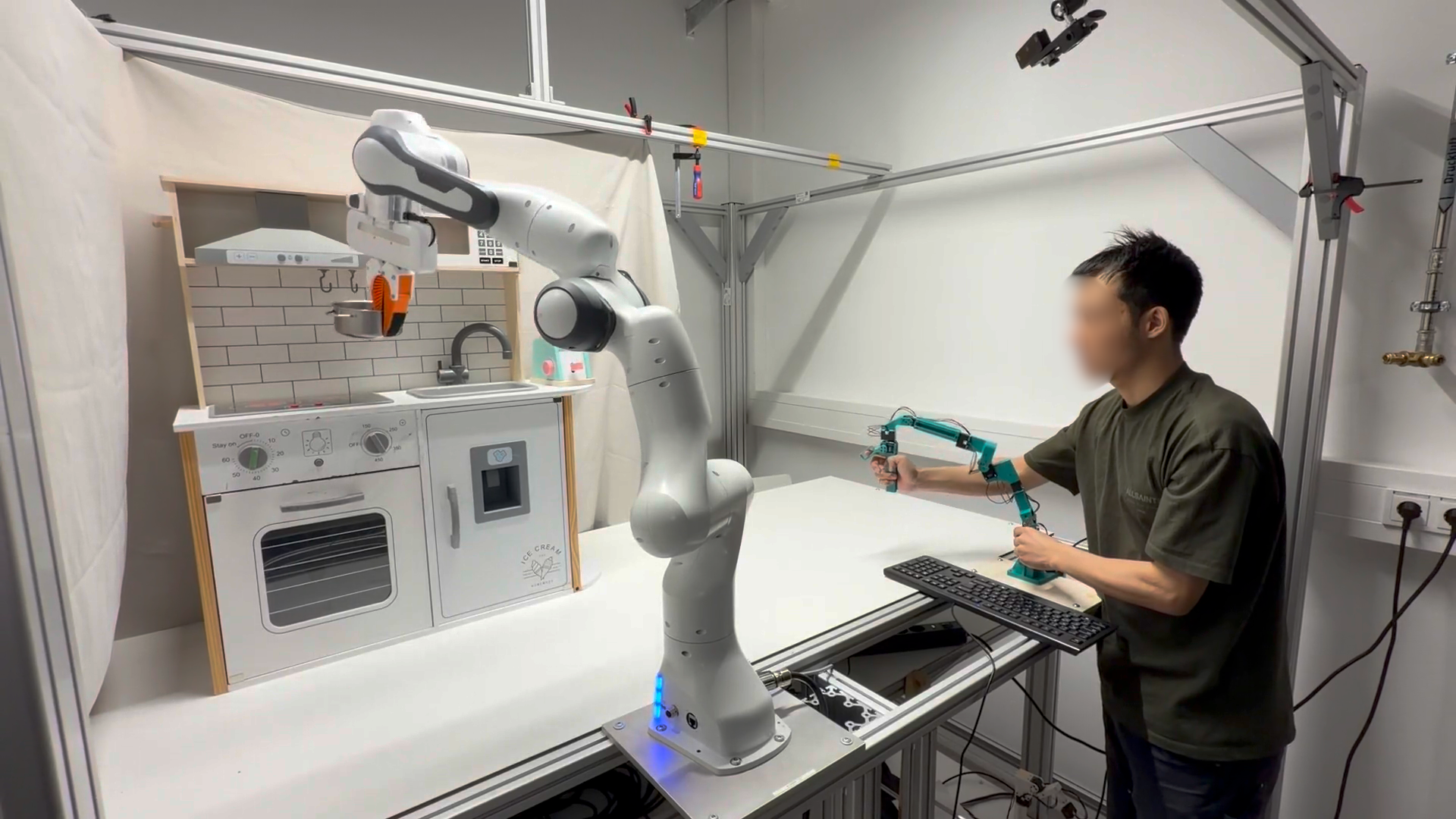}
    \caption{GELLO leader-arm teleoperation.}
    \label{fig:teleop:gello}
  \end{subfigure}
  \caption{\textbf{Two teleoperation modalities currently supported by
  \texttt{nautilus-collect}.} Both devices sit behind the same controller
  abstraction (\S\ref{app:nautilus-collect:datacollect}); the
  data-collection loop is unchanged across modalities.
  \textbf{(\subref{fig:teleop:vr})}
  Oculus Quest 2/3 driving the Franka reference platform in
  relative-pose mode with in-band success/failure labels.
  \textbf{(\subref{fig:teleop:gello})}
  A GELLO kinematically-similar leader arm: the operator's joint
  stream is mapped one-to-one onto the follower, which is preferable
  for contact-rich tasks where the free-hand impedance of a VR
  controller is a poor match.}
  \label{fig:teleop}
\end{figure}

\paragraph{Teleoperation.} Two teleoperation modalities sit behind
the same controller abstraction, so that the data-collection loop is
agnostic to which device is on the operator end. \textbf{VR} drives
Meta Quest 2/3 controllers in relative-pose mode with in-band
episode boundaries and success/failure labels that the operator
triggers mid-recording without leaving the rig --- so the final flag
lands at the same moment the episode ends, removing the post-hoc
relabelling pass other open data-collection pipelines defer to.
\textbf{GELLO} is a kinematically-similar leader arm whose joint
stream is mapped one-to-one onto the follower; we use it for
contact-rich tasks where the free-hand impedance of a held-in-hand
controller is a poor match. Both controllers emit the same
\texttt{Action}, so adding a third modality is a one-class extension.
Figure~\ref{fig:teleop} shows the two rigs side by side.

\paragraph{Synchronised observation.} Per timestep the recorder
snapshots joint state, end-effector pose, gripper state, multi-camera
RGB-D, and a shared ROS-namespace timestamp. A short
approximate-time-synchronisation window plus a small state cache
absorb the jitter between the 50\,Hz state stream and the
15--30\,Hz camera stream without dropping samples.

\paragraph{Companion toolbox.} Around the recording loop,
\texttt{nautilus-collect} absorbs the side concerns of robot learning data
collection so they do not leak into policy code. The argument is
operational: every utility we found ourselves wanting in our own
runs --- TFDS/RLDS~\citep{ramos2021rlds} export under the field
names of the public DROID release~\citep{khazatsky2025droidlargescaleinthewildrobot}
so DROID-trained models load our trajectories without a custom
loader, in-band success/failure labelling so episodes do not need
a post-hoc relabelling pass, replay against the original
\texttt{Robot} object so the controller's behaviour is its own
regression test, and side-by-side visualisers for the raw and
converted forms --- ships with the platform rather than being
rebuilt by each policy author. Hand-eye calibration is the only
recurring manual step; its Charuco-board result is published as
static TF and consumed transparently by the camera reader, so
downstream code never sees calibration as a parameter --- it is
part of the platform, not the policy.

\subsection{Integration boundary with the agent}
\label{app:nautilus-collect:agentboundary}

From above, the Tier-2 WebSocket ``real environment'' of
\S\ref{sec:method:comm} wraps the Gym-compatible adapter; the agent
never imports a ROS\,2 symbol, and sees only the \texttt{Observation}
and \texttt{Action} schema converted from the messages of
\S\ref{app:nautilus-collect:schema}. The two-tier transport boundary is
therefore exactly the boundary between what \texttt{nautilus-collect}
ships and what the agent generates. This is what permits the
robot-skill family of \S\ref{sec:method:agent:skills} to onboard a
new embodiment by writing the four artefacts of the recipe in
\S\ref{app:hardware} --- a \texttt{Robot} subclass, a URDF, a ROS\,2
driver wrapper, and a launch file --- without any of those artefacts
touching the WebSocket layer, the schema, or the agent itself.
Modularity in \texttt{nautilus-collect} is operationally the property that
the new-embodiment patch and the agent are disjoint.

\subsection{Build, dependencies, and reproducibility}
\label{app:nautilus-collect:repro}

The control side depends symbol-for-symbol on
Polymetis~\citep{lin2021polymetis} (vendored as a git submodule
because the upstream repository has been archived) plus libfranka;
the perception side depends on the ZED SDK~\citep{stereolabs_zed_sdk}.
All three are version-pinned at the container layer described in
\S\ref{app:hardware}, so the perception stack and the real-time
stack rebuild independently. The only manual recurring step is the
Charuco calibration of \S\ref{app:nautilus-collect:datacollect}, which has
to be redone whenever a camera is physically moved.

\section{Implementation details supporting the Method}
\label{app:method-impl-details}

This section collects implementation-level details that support
\S\ref{sec:method} but are not necessary for the main conceptual
argument. The main text presents \sysname{} as a substrate
(typed contracts, chambered execution, uniform transport) plus a
content layer (Guides, Sensors, State, and workspace artefacts). The
details below document how that anatomy is realised in the plugin,
how runs are recorded, how registry entries are verified, and how
hardware entries are pinned for reproducibility.

\subsection{Content-layer implementation stack}
\label{app:method-impl:layers}

The content layer is implemented as six layers around a commodity
Claude Code loop. \Lone{} is external to \sysname{}: it provides the
base code-agent loop, file access, shell access, and conversational
state. \sysname{} contributes \Ltwo{}--\Lsix{}. \Ltwo{} exposes
user-facing slash commands and description-driven skills that route a
request either to a deterministic script or to a subagent workflow.
\Lthree{} contains task-specialised subagents for environment probing,
benchmark wrapping, policy serving, training, finetuning, and robot
integration; each subagent runs in an isolated context and the main
thread orchestrates dispatch. \Lfour{} is the tool, validation, and
registry boundary: lifecycle hooks, read-only MCP registry tools,
contract utilities, and WebSocket utilities live here. \Lfive{} is the
domain knowledge library, including the InstallationPlan schema, the
IL-vs-RL protocol, capability-matching rules, smoke contracts, and
observation/action mapping rules. \Lsix{} is the run-record layer that
leaves process logs and user-facing receipts in the target repository.

\begin{table}[t]
\centering\small
\setlength{\tabcolsep}{3pt}
\caption{Six-layer implementation stack of the \sysname{} plugin. \Lone{}
is the commodity agent loop; \Ltwo{}--\Lsix{} route requests, execute
subagents, validate artefacts, read State, encode domain priors, and
leave reproducibility records over the substrate surface.}
\label{tab:layers}
\begin{tabular}{@{}lp{0.22\linewidth}p{0.18\linewidth}p{0.32\linewidth}@{}}
\toprule
Layer & Role in harness & Implements & Contents \\
\midrule
\rowcolor{layerone}
\Lone{}   & Commodity agent loop          & External                    & Claude Code agent loop \\
\rowcolor{layertwo}
\Ltwo{}   & User-facing routing           & Guides                      & Slash commands; skills \\
\rowcolor{layerthree}
\Lthree{} & Task-specialised execution    & Guides                      & Isolated subagents (App.~\ref{app:subagents}) \\
\rowcolor{layerfour}
\Lfour{}  & Validation and State boundary & Sensors / State / substrate & Runtime Sensors; MCP registry; contract \& WebSocket utilities \\
\rowcolor{layerfive}
\Lfive{}  & Domain reference library      & Guides                      & Schemas, classification protocols, validation contracts \\
\rowcolor{layersix}
\Lsix{}   & Run record and hand-off       & Workspace artefacts         & Append-only logs; reproduction receipts \\
\bottomrule
\end{tabular}
\end{table}

These layers map onto the Method concepts as follows. \Ltwo{},
\Lthree{}, and \Lfive{} primarily implement Guides; \Lfour{} implements
runtime Sensors and State access; \Lsix{} implements workspace
artefacts. The important design choice is that robotics-specific
knowledge is not hidden inside model weights or pair-specific glue
code. It is exposed as inspectable schemas, protocols, contracts, and
receipts that can be reviewed and extended independently.

\subsection{Workspace artefacts}
\label{app:method-impl:workspace}

Every \sysname{} run leaves two classes of artefacts in the target
repository. \emph{Process logs}
(\texttt{<repo>/.nautilus/run-log/NN-<task>.md}) are append-only
engineering records. Each names the tool, subagent, command, and
one-line result, so an auditor or re-runner can reconstruct the run in
chronological order. \emph{Receipts} are end-of-run summaries written
to the repository root, such as \texttt{install.md},
\texttt{history.md}, and \texttt{benchmark.md}. They are regenerated
on every run and serve as the hand-off document for a human user:
installation recipe, probe-and-decision narrative, and benchmark
task/protocol guide.

The split is deliberate. Process logs are low-level engineering
evidence, while receipts are concise reproduction instructions. Mixing
the two produces records that satisfy neither audience: too noisy for a
user, too lossy for an auditor.

\subsection{Registry sharing and benchmark verification}
\label{app:method-impl:curation}

Registry mutation is intentionally excluded from the runtime MCP
server. A user can package a policy, benchmark, or robot as an
\texttt{unverified} registry entry with pinned source and environment
artefacts. These entries are already useful for sharing: another user
can reproduce the same setup from the recorded commit, image ID,
contract spec, and run receipt.

\sysname{} maintains \texttt{verified} benchmark entries because
benchmark protocols grow more slowly than policy configurations.
Verification is based on \emph{cross-policy evidence}: for a fixed
benchmark wrapper, multiple released policies---each replayed with its
official checkpoint---should reproduce their published reference
results, with deviations recorded transparently rather than converted
into a universal pass/fail threshold. The verification command fills
reproducibility metadata such as Docker image ID and image size, while
entries that build but lack sufficient cross-policy evidence remain
\texttt{unverified}.

For a verified entry, registry pre-flight can return a short
\texttt{quick\_start} recipe (clone, pull the pinned image, compose up)
instead of re-running the full probe--render--build--classify path. The
registry therefore acts as a shareable cache of reproducible contracts,
not as mutable agent memory.

\subsection{Hardware artefact pinning}
\label{app:method-impl:hardware-pinning}

Hardware entries follow the same verified-registry pattern as policies
and benchmarks. Each hardware entry pins three artefacts: the
\texttt{nautilus-collect} source commit implementing the Robot
contract, the corresponding \texttt{nautilus-collect} Docker image,
and a platform-specific driver-stack image. The driver-stack image owns
the vendor-side dependency closure, such as Franka's
\texttt{libfranka} plus Polymetis, UFactory xArm's Python SDK, the
Allegro Hand ROS\,2 driver package, or Unitree H1's
\texttt{unitree\_sdk2} plus \texttt{unitree\_ros2}. The platform
container above the driver stack consumes only the standard Robot
contract and the Tier-2 policy endpoint, so policy-side code is not
rewritten when the hardware container changes.

This is the hardware analogue of the benchmark and policy registry:
the agent looks up a verified contract and pinned artefacts rather than
hand-wiring a driver stack from scratch. Adding a new embodiment is
therefore framed as implementing and verifying a Robot contract entry,
not as modifying policy or benchmark code.

\subsection{Surgical-pruning}
\label{app:method-impl:surgical-pruning}

For each released tuple $(P_i, B_j, c_{ij})$, where $P_i$ is a public policy repository,
$B_j$ is a target benchmark,
and $c_{ij}$ is the released checkpoint used in the reference result,
we construct a pruned repository $P_i^{\setminus B_j}$.
We developed an additional agentic plugin to automatically prune the target-benchmark-specific support from $P_i$:
example directories, benchmark adapters, task configs, dispatch entries,
benchmark-specific scripts, documentation references, and other code paths whose purpose is to run $P_i$ on $B_j$.
The pruning is surgical rather than destructive:
supports of other non-target benchmarks in the same repository $P_i^{\setminus B_j}$ must remain usable after pruning, so the integration test starts from a healthy policy repository with only the target integration missing.
To make checkpoint loading possible without leaking $B_j$ integration code, we leave a short checkpoint note in $P_i^{\setminus B_j}$ containing only high-level natural language descriptions of the non-trivial policy-specific facts when loading the checkpoint $c_{ij}$.

\section{Subagent walkthroughs}
\label{app:subagents}
\label{app:benchmark-skill}
\label{app:skill-walkthroughs}

This appendix supports \S\ref{sec:method:agent:skills} and
\S\ref{sec:exp:reproduction}. It is organised around
Table~\ref{tab:subagents}: each L3 subagent gets its own
subsection, followed by the cross-evaluation \emph{orchestrator}
that composes them. We flag the distinction up front:
cross-evaluation (\S\ref{app:subagents:crosseval}) is not itself a
subagent --- it is the workflow that wires
\texttt{policy-generator}, \texttt{benchmark-generator}, and the
State registry together via the \nautcmd{/nautilus:eval} slash
command. All subagents otherwise share the same recipe ---
\emph{probe} the target repository, \emph{render} a small set of
templated files that wire it into the substrate of
\S\ref{sec:method:platform}, and \emph{smoke} the result against a
fail-fast contract --- and differ only in which substrate invariant
they target and which artefacts they emit.
Table~\ref{tab:slash-commands} lists every slash command;
Table~\ref{tab:subagents} lists every subagent.

\begin{table}[t]
\centering\small
\caption{L2 slash commands exposed by the \sysname{} plugin.}
\label{tab:slash-commands}
\begin{tabular}{@{}lp{0.65\linewidth}@{}}
\toprule
Command & Purpose \\
\midrule
\nautcmd{/nautilus:help}      & print a user-facing overview of the plugin \\
\nautcmd{/nautilus:benchmark} & browse, submit, or verify benchmark registry entries \\
\nautcmd{/nautilus:policy}    & browse, submit, or verify policy registry entries \\
\nautcmd{/nautilus:robot}     & browse robot registry entries \\
\nautcmd{/nautilus:eval}      & end-to-end cross-evaluation: clone both repos, build containers in parallel, render an obs/action adapter, run a one-episode smoke \\
\nautcmd{/nautilus:train}     & train a registered policy on a benchmark-native dataset (Mode~A) or a LeRobot Hugging Face dataset (Mode~B); 1-epoch smoke + detached full run \\
\bottomrule
\end{tabular}
\end{table}

\begin{table}[t]
\centering\small
\caption{L3 subagents. Each runs in an isolated context; the main thread orchestrates dispatch.}
\label{tab:subagents}
\begin{tabular}{@{}lp{0.62\linewidth}@{}}
\toprule
Subagent & Role \\
\midrule
\texttt{env-generator}       & probe a target repo, render base \texttt{docker/}, classify \\
\texttt{benchmark-generator} & extend an existing image for a sim benchmark; IL/RL classification, scaffolding, smoke test \\
\texttt{policy-generator}    & wrap a checkpoint as a WebSocket server with multi-benchmark dispatch \\
\texttt{train-generator}     & cross-pollinate a policy with a benchmark-native or LeRobot dataset; render mapper, training script, 1-epoch smoke \\
\texttt{model-finetune}      & LoRA, freeze-backbone, or full-finetune scripts in PyTorch or JAX/Flax \\
\texttt{robot-generator}     & interface-defined stub; full implementation deferred \\
\bottomrule
\end{tabular}
\end{table}

\subsection{\texttt{env-generator}: base scaffolding and idempotent extension}
\label{app:subagents:env-generator}
\label{app:benchmark-skill:arch}
\label{app:benchmark-skill:extend}

The claim of \texttt{env-generator} is that
\emph{every other subagent can be written without knowing how the
target repository was built}. It establishes this by reducing a
heterogeneous source repository to a small file-system contract
(the \texttt{docker/} directory plus a \texttt{.classification}
label) that downstream subagents read but never re-derive. The
substrate it pins is the one most likely to bite a generic harness:
target repositories in this domain mix system libraries (Vulkan,
MuJoCo, ROS), C++ extensions compiled against a specific CUDA
toolchain, and Python packages with sharp version constraints, and
no single package manager covers all three. We pair Docker with
\texttt{uv} --- Docker for the system and toolchain layer,
\texttt{uv} for a single-pass Python resolution with reproducible
lock files --- and reject \texttt{conda}/\texttt{poetry}/\texttt{pipenv}
on entry; the result is one version-controlled artefact that owns
all three layers, which is what makes downstream composition safe.

Two design choices keep the subagents structurally decoupled.
\emph{First}, downstream subagents are coupled to
\texttt{env-generator} only through three file-system contracts
--- the \texttt{docker/} layout, the \texttt{.classification} label
($\in\{$\texttt{plain}, \texttt{benchmark}, \texttt{policy},
\texttt{robot}$\}$), and four named anchor comments embedded in
the base templates --- never through shared in-memory state. Every
intermediate result is therefore independently inspectable and the
pipeline can resume from any checkpoint, which matters when an
agent run is interrupted at the boundary between two subagents.
\emph{Second}, every injection is wrapped in sentinel-guarded
blocks of the form
\begin{verbatim}
# --- NAUTILUS OPEN: needs-vulkan ---
...
# --- NAUTILUS CLOSE: needs-vulkan ---
\end{verbatim}
\noindent and the helper that performs an injection skips it
whenever the opening sentinel already exists. This is what makes
re-running any downstream subagent idempotent --- a property we
rely on whenever a smoke test fails partway through and the user
edits and retries.

\subsection{\texttt{benchmark-generator}: extending the base for sim benchmarks}
\label{app:subagents:benchmark-generator}

Sim benchmarks fail in three structurally distinct layers --- the
container build, the reward function, and the policy--environment
interface --- and a generic end-to-end smoke conflates them, so the
agent that has to retry on failure cannot tell which scaffold is
broken. \texttt{benchmark-generator} is organised around catching
each layer separately, which in turn forces an IL/RL split because
the L3 probes for the two regimes share nothing.
\label{app:benchmark-skill:ilrl}\label{app:benchmark-skill:smoke}\label{app:benchmark-skill:artefacts}

The IL/RL classifier uses a single primary criterion --- whether
\texttt{step()} returns a meaningful reward signal --- and falls
back to secondary cues (policy architecture, PPO/SAC/TD3 imports,
presence of offline datasets) only on borderline cases such as
benchmarks that expose a reward function for compatibility but are
conventionally evaluated under IL. RL targets are further split
into a \emph{dispatcher} mode (the repository's own training
script is wrapped) and an \emph{sb3} mode (a Hydra-driven
\texttt{stable-baselines3} scaffold). The split is load-bearing
because IL and RL targets demand different evaluation entry points
and different smoke criteria; collapsing them would force every
target into a least-common-denominator rollout that fails to catch
either family's typical errors.

The smoke contract runs four checks in sequence and halts at the
first failure, surfacing a typed error to the Sensor layer of
\S\ref{sec:method:sensors}. The argument for layering, rather than
a flat end-to-end rollout, is diagnostic: an L1 failure implicates
the build, L2 the reward function, and L3 the policy--environment
interface, so the level at which the contract breaks tells the
agent which scaffold to retry. L3-IL and L3-RL are mutually
exclusive; the classification picks one.

\begin{center}
\small
\begin{tabular}{@{}p{0.07\linewidth}p{0.30\linewidth}p{0.55\linewidth}@{}}
\toprule
\textbf{Level} & \textbf{Test} & \textbf{Pass criterion} \\
\midrule
L1    & \texttt{env.reset()} + 10 random steps & no exception; observation shape matches spec \\
L2    & 100-step rollout with reward logging   & reward finite and non-constant \\
L3-IL & WebSocket handshake with a random-action policy server & client--server round-trip completes \\
L3-RL & 100-step training run                   & loss finite and decreasing \\
\bottomrule
\end{tabular}
\end{center}

\subsection{\texttt{policy-generator}: wrapping a checkpoint as a WebSocket server}
\label{app:subagents:policy-generator}
\label{app:skill-walkthroughs:policy}

Wrapping an inference-only repository as a \emph{policy container}
(\S\ref{sec:method:agent:sandbox}) has two problems that the
WebSocket layer of \S\ref{app:interfaces:protocol} does not solve:
locating an inference entry point in a repository that may not
document one, and providing a smoke contract for VLA-class
checkpoints too large to host on the scaffolding machine.
\texttt{policy-generator}'s design lives entirely in those two
corners. For the first, a scanner pairs candidate classes (those
defining \texttt{forward}/\texttt{predict}/\texttt{infer}) with
checkpoint loaders and cross-references the README; when heuristics
disagree, it falls back to an interactive prompt rather than
guessing, on the principle that a wrong entry point is worse than
a paused agent. Once located, chunked outputs of shape $(H,
\text{act\_dim})$ with $H>1$ are auto-wrapped behind an
\texttt{ActionChunkBroker} so chunk-prediction policies serve the
same per-step protocol as one-step policies and the benchmark side
stays untouched.

For the second, the smoke contract admits three modes ---
\texttt{with\_ckpt}, \texttt{mock\_forward}, and \texttt{skipped}
--- and the middle one is load-bearing. If smoke required a real
checkpoint, the contract would silently become opt-out for exactly
the policies that are tens of gigabytes and most likely to expose
integration failures. \texttt{mock\_forward} monkey-patches
\texttt{infer} to return a random action tensor of the declared
shape and still exercises the full container lifecycle (start,
healthz, accept obs, return action), so the WebSocket plumbing is
verified even when the weights are absent.

\subsection{\texttt{train-generator}: cross-pollinating policies with datasets}
\label{app:subagents:train-generator}
\label{app:skill-walkthroughs:training}

Cross-pollinating a wrapped policy with someone else's dataset
would normally require hand-tuning three things at once: the
dataloader to the policy's training-time constructor, the dataset's
gripper sign convention to the benchmark's, and the gap between the
inference-only and training-time constructors of the policy class.
\texttt{train-generator} avoids all three by running introspection
inside the \emph{running} policy container --- the chambered
execution surface of \S\ref{sec:method:platform} --- rather than
in the agent's main thread, and reusing the cross-evaluation
mapping rules at data-loading time. Three probes there read the
dataloader's obs/action schema, the training-side hyperparameters
from the policy repository (with README defaults as fallback), and
the training-time constructor signature. The dataset and policy
specs are then fed through the same five mapping rules that
cross-evaluation uses (\S\ref{app:subagents:crosseval}), but
applied at data-loading time, so the inference-time adapter and
the training-time mapper share their generator.

Two dataset modes cover the empirical access patterns we have seen.
\emph{Mode~A} bind-mounts the benchmark's native dataset
into the container and reuses the benchmark's own dataloader,
which preserves benchmark-specific conventions such as gripper
sign. \emph{Mode~B} installs \texttt{lerobot} inside the container
and pulls a Hugging Face dataset, unlocking the LeRobot ecosystem
at the cost of a Python $\geq 3.12$ requirement that the
spec-capture script pre-checks before any rebuild. The choice is
made by the \texttt{benchmark=}/\texttt{lerobot=} flag on
\nautcmd{/nautilus:train}. The training entry point is a thin
wrapper that imports the policy with probed constructor arguments,
injects the mapper as a data transform, and calls the policy's own
\texttt{train} method rather than reinventing a training loop. The
1-epoch smoke is capped at ${\sim}8$ steps and 30 minutes, on the
same diagnostic principle as the benchmark smoke ladder: a short
run that verifies loss is finite and decreasing, that a checkpoint
was written, and that GPU memory stayed in budget is enough to
catch the integration failures that would otherwise surface 12
hours into a full run.

\subsection{Cross-evaluation: orchestrator across subagents}
\label{app:subagents:crosseval}
\label{app:skill-walkthroughs:crosseval}

Cross-evaluation is the workflow that operationalises the
$\Theta(N\!+\!M)$ collapse claim of \S\ref{sec:method:state}: it
takes any two registry entries and produces a running rollout
without writing pair-specific glue. The defining choice is that
compatibility is decided \emph{before} any container starts, by
comparing the two registry specs (obs keys, action dim, gripper
convention) and bucketing the pair as \emph{native} (specs match,
run directly), \emph{compatible-zero-shot} (structurally
compatible after a few well-defined transformations, render an
adapter), or \emph{incompatible-action} (fundamentally
mismatched, abort with a typed error pointing to retraining).
The argument for this pre-render gate is operational: a runtime
mismatch between policy and benchmark surfaces silently as a
\texttt{ConnectionRefused} or a shape error well after expensive
containers have been brought up, whereas the spec-level check
fails in seconds and tells the agent which side to fix. Around
this gate, \nautcmd{/nautilus:eval} resolves names through MCP
\texttt{lookup\_policy}/\texttt{lookup\_benchmark}
(\S\ref{app:registry}), short-circuits already-verified entries to
their published images, otherwise dispatches \texttt{env-generator}
plus the matching downstream subagent in parallel on each side, and
finishes with a one-episode smoke that emits
\texttt{cross-eval-report.md}.

For the \emph{compatible-zero-shot} bucket, the adapter is
auto-generated from the two spec JSONs using five rules in a fixed
order: \emph{key rename} (e.g.\ \texttt{agentview\_rgb} $\to$
\texttt{image}), \emph{chunk split}, \emph{dim slice}, \emph{dim
pad} (zero-pad trailing dims when the policy expects a wider input
than the benchmark exposes), and \emph{image preprocess} (\texttt{uint8}
HWC $\to$ \texttt{float32} CHW with normalisation). The adapter is
injected as a preprocessing step inside the policy server's
\texttt{infer}, leaving the benchmark side untouched, which
preserves the property that the benchmark image we cache is the
same image any other policy will be evaluated against. As a worked
example, \nautcmd{/nautilus:eval policy=openpi-pi0 benchmark=libero}
classifies as \emph{compatible-zero-shot} (LIBERO exposes
\texttt{agentview\_rgb} where OpenPI-$\pi_0$ expects \texttt{image};
action dims match), renders an adapter that combines a key rename
with image preprocess, and runs one LIBERO episode --- the only
artefact the user sees is a Markdown report recording the bucket,
the applied rules, the rollout outcome, and per-step inference
latency from the WebSocket protocol's \texttt{server\_timing}
field.

\section{Typed registry and State layer}
\label{app:registry}

This appendix expands \S\ref{sec:method:state} into the schema,
tools, curation protocol, and matching logic that together
realise the State component of harness engineering at the
robotics-research scale. The argument back to
\S\ref{sec:method:state} is that the registry is a
\emph{contract carrier}: it stores enough machine-introspectable
metadata that any policy and any benchmark can be composed without
hand-written adapters, but it stores no code and no agent state,
so its mutation surface is small and human-auditable.

\subsection{Schema and MCP tools}
\label{app:registry:schema}

The registry is split into an \emph{index} layer that lists what
is registered and a \emph{spec} layer that records the per-entry
contract metadata an agent needs to compose two entries (obs/action
shapes and dtypes, reward structure for benchmarks, released
checkpoint URLs for policies, optional training block for
datasets). The split is what makes the registry small enough to
audit during review: the index changes whenever someone adds an
entry, but the spec for a given entry is rewritten only when its
on-disk contract genuinely changes. JSON-schema validation runs at
write time so a malformed entry never lands in the index.

Agents read the registry through a FastMCP server rather than the
filesystem, and the server exposes only read-only tools. Two
properties follow. First, the on-disk format can evolve without
breaking agent code, since stability is guaranteed at the tool
return shape, not at the YAML/JSON layout. Second --- and this is
why the registry is suited to be a contract carrier rather than a
mutable store --- there is no write API at runtime; the only way
to mutate the verified registry is through a reviewed registry update.

\subsection{Curation and fuzzy matching}
\label{app:registry:curation}

The registry flow of \S\ref{sec:method:curation}
operationalises the boundary between the registry as
machine-introspected data and the registry as human-attested
record. A user's \emph{submission} runs local pre-flights
(clean git tree, Docker login, no name or URL collision in the
index) and appends an \texttt{unverified} entry, deferring to the
user to share the generated evidence package. A maintainer's
\emph{verification} checks reproducibility pins, re-runs the relevant
smoke contracts inside the proposed container, and marks benchmark
entries as \texttt{verified} only when independent evidence supports
the same conclusion. The MCP server is deliberately read-only, so
runtime agents cannot modify verified entries.

Lookup is tolerant by design, because users refer to repositories
by names that the registry should not have to canonicalise on
their behalf. A four-tier policy is applied in order ---
exact-name, exact-URL, URL-basename, and an aggressively
normalised key match (lowercased, trailing digits stripped,
non-alphanumerics removed) --- and the matched tier is returned
alongside the entry so downstream code can prefer high-confidence
matches when disambiguation is needed. In practice
\texttt{lookup\_benchmark("maniskill")},
\texttt{lookup\_benchmark("ManiSkill")}, and a fork URL all reach
the same entry; what we surface to the agent is the tier at which
each match landed.

\subsection{Hardware registry parallel}
\label{app:registry:hardware}

Hardware platforms enter the registry through the same
submission--verification flow as benchmarks and policies: the four
currently registered platforms --- Franka Panda (single-arm and
bimanual), UFactory xArm, Allegro Hand, and Unitree H1 of
Table~\ref{tab:hardware:platforms} --- each is represented by pinned
source and container artefacts, with verification confirming that the
hardware container builds and that the upward ROS\,2 schema of
\S\ref{app:nautilus-collect:schema} matches. The argument back to
\S\ref{sec:method:hardware} is straightforward: from the agent's
perspective the four-method \texttt{Robot} contract is what makes
hardware addressable, and from the registry's perspective hardware
carries no special-case logic, so a fifth platform enters
\sysname{} along the same path as a fifth benchmark or a fifth
policy.

\section{Runtime sensor reference}
\label{app:sensors}

This appendix is the long-form reference for the six runtime
Sensors of \S\ref{sec:method:sensors}, ordered earliest (before any
tool runs) to latest (after deployment). A separate curation check
(human attestation at registry promotion) lives at the curation
boundary and is described in \S\ref{sec:method:curation}.
Each Sensor is a narrow guardrail around one boundary where agent
mistakes otherwise become expensive: shell execution, rendered
configuration, captured interfaces, policy--benchmark composition,
deployment smoke tests, and post-run cross-run verification. The common
contract is that a Sensor does not add new task logic; it checks a
machine-readable artefact or run result, then either lets the workflow
continue, asks the agent to repair the artefact locally, or blocks the
workflow until a human or registry update resolves the mismatch.

\begin{table}[h]
\centering\small
\setlength{\tabcolsep}{3pt}
\caption{The six runtime Sensors fired by \sysname{}, ordered earliest to latest in the workflow. Registry verification is described in \S\ref{sec:method:curation}.}
\label{tab:sensors}
\begin{tabular}{@{}p{0.18\linewidth}p{0.24\linewidth}p{0.31\linewidth}p{0.21\linewidth}@{}}
\toprule
Sensor & Fires when & Checks & Response \\
\midrule
Pre-action filter & before any tool call, via Claude Code's PreToolUse hook & Shell commands for destructive patterns such as \texttt{rm -rf /}, fork bombs, \texttt{mkfs} on a device, or \texttt{docker volume prune --force}. & Block the tool call and require a safer command. \\
Render-time auditor & before writing generated \texttt{docker-compose} files & Whether private registry URLs or verified images leak into the from-scratch reproduction path, where the default must remain rebuildable by the user. & Rewrite the rendered file or stop before the file is committed to the workspace. \\
Interface verify & during benchmark or robot capture & Whether observation, proprioception, and action specs are task-invariant and match the typed contract expected by \S\ref{sec:method:interfaces}. & Reject the captured spec and ask the agent to normalise the wrapper. \\
Spec-comparison gate & before any heavyweight run; registry-only, no docker & The policy spec against the benchmark or robot spec, including observation keys, dtypes, action shape, and control mode. The gate reports the native, compatible-zero-shot, or incompatible-action bucket. & Proceed, request an adapter, or block the run before container construction. \\
Tiered smoke ladder & at deployment, before scoring & Increasingly realistic execution checks: endpoint liveness, \texttt{reset}/\texttt{step}, finite observations and rewards, and IL/RL-specific gating such as a 1-epoch training smoke. & Keep repair local to the failing tier; only promote to scoring after the tier passes. \\
Cross-run verification & post-run, before verified benchmark status & Whether independent runs of the same benchmark wrapper support the same conclusion despite variation from seeds, checkpoints, and hardware. & Keep insufficiently supported entries \texttt{unverified} and record the evidence for later reproduction. \\
\bottomrule
\end{tabular}
\end{table}

For the $\pi_0$--LIBERO trace in Figure~\ref{fig:method-trace}, the
spec-comparison gate is the first Sensor that uses both sides of the
registry: it reads the generated $\pi_0$ Policy contract and the LIBERO
Benchmark/Environment contract, then decides whether the pair can be run
natively, needs an observation/action adapter, or should be blocked as
incompatible. The later smoke ladder tests the same decision against a
live WebSocket endpoint, and cross-run verification asks whether the
benchmark wrapper has enough independent evidence to become a verified
registry entry.

\section{Real-robot deployment details}
\label{app:hardware:deployment-details}

This appendix expands the compact evidence block of
\S\ref{sec:exp:hardware}. The main paper uses the hardware examples
as a systems demonstration: unchanged policy code, unchanged WebSocket
envelope, unchanged platform-side container, and a hardware-container
swap. Here we spell out the two deployments: a VLA manipulation
checkpoint on Franka and an RL-trained locomotion policy on H1, plus
the collect--fine-tune--redeploy loop available when real-world
adaptation is needed.

\paragraph{Franka manipulation setup.}
We take the $\pi_{0.5}$ checkpoint reproduced in
\S\ref{sec:exp:reproduction} (whose wrapper score against LIBERO is
reported in Table~\ref{tab:trust_validation}) and deploy it on a
Franka Panda single-arm manipulator. The platform-side container is
unchanged from the simulation rollout; the only swap is the
hardware container of \S\ref{app:hardware}, released alongside the
platform.

\paragraph{From benchmark wrapper to real-robot rollout.}
After the $\pi_{0.5}$ checkpoint is wrapped for benchmark evaluation,
the same policy artefact is launched against the Franka hardware
container. The launch file and policy artefact are unchanged across
the simulation-to-real boundary, and no policy-code edits are required.

\paragraph{Fine-tuning recovery.}
The released $\pi_{0.5}$ checkpoint does not zero-shot complete
pick-and-place in our real setup. When adaptation is needed,
performance recovery uses the same \sysname{} workflow: collect
demonstrations through \S\ref{app:nautilus-collect:datacollect},
fine-tune the checkpoint, and redeploy through the same launch path.
The important systems point is that data collection and deployment
share the same code path below the action source: the human
teleoperator and policy server both call into the same \texttt{Robot}
interface, with only the actor producing \texttt{apply\_action}
changing.

\paragraph{Unitree H1 locomotion setup.}
We deploy an RL-trained locomotion policy on a Unitree H1 humanoid.
Unlike the Franka case, this is not a VLA manipulation checkpoint;
it is a different policy family and a different control regime. The
deployment nevertheless uses the same policy-facing WebSocket
endpoint and the same platform-side container that drove the Franka
deployment; only the H1 hardware container and launch namespace
change.

\paragraph{H1 deployment.}
The H1 example is not intended as a state-of-the-art locomotion
benchmark. Its purpose is to show that the same deployment substrate
that drove a single-arm VLA manipulation rollout also supports an
RL-trained legged-locomotion rollout without changes to the platform
container, WebSocket envelope, or policy-side integration code.

\section{Full Trust Verification Experiment}
\label{app:full_trust_verification}

Table~\ref{tab:benchmark_comparison} summarizes our full trust verification
study across five trust-validated robot learning benchmarks. Each entry reports
our reproduced score followed by the reference score reported by the
corresponding official paper, repository, or benchmark leaderboard. Unless
otherwise stated, all scores are success rates in percentage points. We use the
officially released checkpoints whenever available; otherwise, we follow the
official training and evaluation protocol and report the result from the
corresponding reproduced checkpoint.

\paragraph{LIBERO.}
For LIBERO, we evaluate on the four standard suites: LIBERO-Spatial,
LIBERO-Object, LIBERO-Goal, and LIBERO-Long. The reference numbers for
Diffusion Policy and OpenVLA are taken from the OpenVLA LIBERO simulation
benchmark. OpenVLA reports results averaged over three random seeds, with
500 rollouts per seed, corresponding to 10 tasks and 50 rollouts per task. The
OpenVLA checkpoints are the official LoRA-finetuned checkpoints for the four
LIBERO suites, with LoRA rank 32; the evaluation uses center cropping because
random crop augmentation was used during fine-tuning. For $\pi_{0.5}$, we use
the official OpenPI LIBERO fine-tuning setup and checkpoint
\texttt{pi05\_libero}; the official OpenPI repository provides the LIBERO
fine-tuning workflow by converting LIBERO data to LeRobot format, computing
normalization statistics, and launching the \texttt{pi05\_libero} training
configuration. For SmolVLA, we follow the official SmolVLA simulation
fine-tuning setup, where only the action expert is trained while the VLM
backbone is frozen; the paper reports 100k fine-tuning steps with batch size
64 for simulation benchmarks and uses 10 flow-matching inference steps.

\paragraph{RoboCasa.}
For RoboCasa, we follow the RoboCasa365 multi-task learning benchmark. The
benchmark trains language-conditioned policies on the human pretraining split,
which contains 300 tasks, including 65 atomic tasks and 235 composite tasks,
with 100 demonstrations per task. Evaluation is performed on the
\texttt{atomic\_seen}, \texttt{composite\_seen}, and
\texttt{composite\_unseen} splits in pretraining scenes. The official
Diffusion Policy baseline uses batch size 192 and is trained for 250k steps.
The official OpenPI $\pi_0$ and $\pi_{0.5}$ baselines use full fine-tuning
with batch size 64 and are trained for 75k steps. The RoboCasa365 simulator
runs at 20 Hz and uses an operational-space controller with a 12-dimensional
action space: seven dimensions for the arm, i.e., 3D translation, 3D rotation,
and gripper control, plus five dimensions for mobile base translation,
rotation, torso height, and mobile-base control gating. We note that the
Diffusion Policy reference value of 15.7 corresponds to the Atomic-Seen split,
whereas the official overall average for Diffusion Policy is 6.1. In contrast,
the $\pi_0$ and $\pi_{0.5}$ values in Table~\ref{tab:benchmark_comparison}
use the official average across the three RoboCasa365 evaluation splits.

\paragraph{ManiSkill.}
For ManiSkill, we follow the simulation benchmark released with the RDT
repository. The benchmark evaluates five ManiSkill tasks:
\texttt{PegInsertionSide}, \texttt{PickCube}, \texttt{StackCube},
\texttt{PlugCharger}, and \texttt{PushCube}. The official protocol generates
5,000 demonstration trajectories using motion planning. For OpenVLA and Octo,
the original absolute joint-position-control trajectories are converted to
delta end-effector pose control to match their action spaces; for RDT and
Diffusion Policy, joint-position-control data are used. Diffusion Policy is
trained from scratch for 1000 epochs, and the epoch-700 checkpoint is selected
based on the lowest validation sample loss. OpenVLA is fine-tuned from the
official pretrained checkpoint using LoRA rank 32. Each method is evaluated
over 250 trials, consisting of 10 random seeds and 25 trials per seed.

\paragraph{RoboTwin.}
For RoboTwin, we report Easy and Hard success rates separately rather than
mean and standard deviation. RoboTwin 2.0 evaluates 50 dual-arm manipulation
tasks. All models are trained on 50 clean, non-randomized demonstrations per
task. The Easy setting evaluates policies in clean environments, while the
Hard setting evaluates the same policies under domain-randomized environments
with clutter, lighting variation, table-height variation, background textures,
and unseen language instructions. ACT is trained with chunk size 50, batch
size 8, single-GPU training for 6,000 epochs, and temporal aggregation during
deployment. Diffusion Policy is trained for 600 epochs with batch size 128
and planning horizon 8. $\pi_0$ is fine-tuned for 30,000 steps with batch
size 32. We therefore report RoboTwin as
``Easy success / Hard success'' rather than using a $\pm$ notation.

\paragraph{ALOHA.}
For ALOHA, the reference scores are taken from the OpenVLA-OFT ALOHA
experiments. These experiments use a bimanual ALOHA platform with two ViperX
300 S arms, three camera views, 14-dimensional joint-angle robot state, and
target absolute joint-angle actions at 25 Hz. The ALOHA suite contains four
real-world tasks: \texttt{fold shorts}, \texttt{fold shirt},
\texttt{scoop X into bowl}, and \texttt{put X into pot}. The demonstrations
per task are 20, 30, 45, and 300 respectively, and the evaluation uses 10,
10, 12, and 24 trials respectively. The ALOHA metric is an aggregate task
performance score based on a predefined partial-completion rubric rather than
a single binary success indicator for all tasks. ACT and Diffusion Policy are
trained from scratch on each task. ACT uses action chunks of 25 steps and is
trained for 10k--70k epochs depending on the task. Diffusion Policy uses a
2-step observation history, predicts 24-step action chunks, and is trained for
30k--120k steps depending on the task. The $\pi_0$ reference is obtained by
fine-tuning the official model with the default full fine-tuning recipe and
training until convergence.

\begin{figure}[t]
\centering
\includegraphics[width=\linewidth]{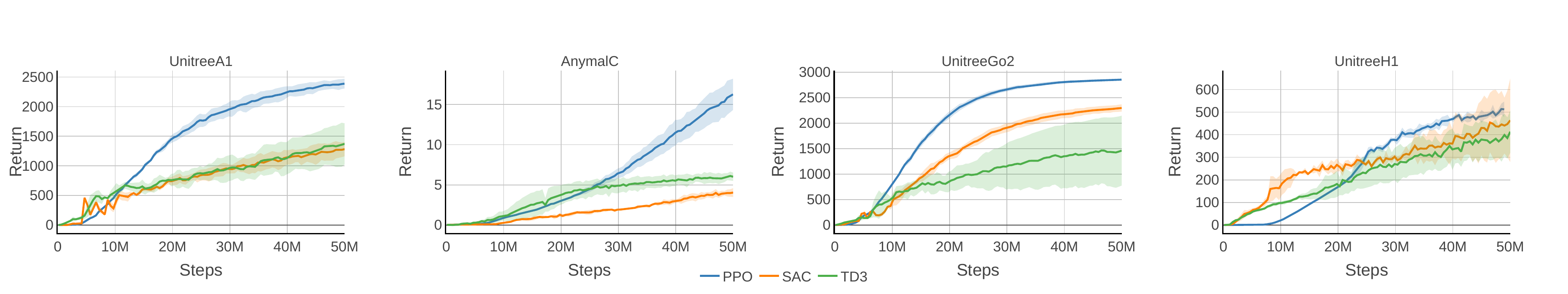}
\caption{The performance benchmark on Loco-Mujoco with three classical RL algorithms, including PPO, SAC, and TD3.}
\label{fig:loco-mujoco}
\end{figure}

\paragraph{Loco-Mujoco (RL)} 
We conduct additional experiments on Loco-Mujoco~\citep{alhafez2023b}, an RL locomotion benchmark implemented in JAX that includes both humanoid and quadrupedal robots. We evaluate three classical algorithms: PPO~\citep{schulman2017proximal}, SAC~\citep{haarnoja2018soft}, and TD3~\citep{fujimoto2018addressing}. Since Loco-Mujoco only provides a native implementation of PPO, we further incorporate SAC and TD3 from Stable-Baselines3 and reimplement them in JAX to demonstrate the adaptability of our framework. As shown in Fig.~\ref{fig:loco-mujoco}, all three algorithms successfully acquire locomotion skills. Moreover, for H1, we successfully transfer the learned policy to the real world.

\paragraph{Comparability and limitations.}
The goal of this experiment is to verify whether our reproduced results match
the official reference numbers under the closest available protocol, rather
than to claim a perfectly controlled cross-benchmark comparison. Several
benchmarks differ in evaluation budget, action representation, robot
embodiment, and metric definition. In particular, RoboTwin reports Easy and
Hard success rates, RoboCasa reports both split-level and average success
rates, and ALOHA reports rubric-based aggregate task performance. We therefore
treat each benchmark independently and compare our reproduction only against
the official protocol used for that benchmark.

\begin{table}[h]
  \centering
  \small
  \caption{\textbf{Benchmark Comparison.}
  Each cell reports our reproduced success rate / the reference success rate
  reported in the original paper. For RoboTwin, values are written as
  \emph{easy (hard)}, where the number outside the parentheses is the
  easy-task success rate and the number in parentheses is the
  hard-task success rate.}
  \label{tab:benchmark_comparison}
  \setlength{\tabcolsep}{4.5pt}
  \begin{tabular}{l ccccc}
    \toprule
    & \multicolumn{5}{c}{Benchmark} \\
    \cmidrule(lr){2-6}
    Method
      & LIBERO
      & RoboCasa
      & ManiSkill
      & RoboTwin
      & ALOHA \\
    \midrule
    Diffusion Policy
      & 70.2 / 72.4
      & 15 / 15.7
      & 32.4 / 30.2
      & 26.4 (0.42) / 28.04 (0.64)
      & 75.8 / 77.5 \\

    ACT
      & -- / --
      & -- / --
      & -- / --
      & 30.0 (1.64) / 29.74 (1.74)
      & 72.8 / 72.3 \\

    $\pi_0$
      & 93.6 / 94.2
      & 15.2 / 14.8
      & -- / --
      & 42.64 (18.62) / 46.42 (16.34)
      & -- / 83.9 \\

    $\pi_{0.5}$
      & 97.0 / 96.85
      & 18.6 / 16.9
      & -- / --
      & -- / --
      & -- / -- \\

    OpenVLA
      & 78.2 / 76.5
      & -- / --
      & 4.0 / 4.8
      & -- / --
      & -- / -- \\

    SmolVLA
      & 88.2 / 87.3
      & -- / --
      & -- / --
      & -- / --
      & -- / -- \\
    \bottomrule
  \end{tabular}
\end{table}

\section{Full agentic-ablation tables}
\label{app:ablation-tables}

This appendix gives the per-pair breakdown of the ablation summarised
in Table~\ref{tab:ablation_summary}. Each of the three pruned
policy--benchmark pairs forms a separate row group, and within each
group the four settings (vanilla coding agent, full \sysname{}, w/o
template, w/o verified image) are evaluated under the same column
schema as the main table: \textbf{wrap success} and \textbf{task
success} (the Pass rate of our metrics), \textbf{agent inference
time}, \textbf{interaction budget} (total agent turns),
\textbf{token/context budget} (total token footprint and peak
context length), and \textbf{cost} in metered USD. Each cell aggregates
$10$ independent trials per setting; time, interaction, token,
context, and cost columns are averaged over accepted trials.

\begin{table}[t]
  \centering
  \small
  \caption{\textbf{Per-pair ablation breakdown (Claude Opus 4.7 /
  Claude Sonnet 4.6).}
  Same column schema as Table~\ref{tab:ablation_summary}.
  Each pair is denoted
  $\mathrm{Policy}^{\setminus\,\mathrm{Benchmark}}$, where the
  superscript $\setminus\,\mathrm{Bench}$ marks the benchmark whose
  support has been surgically removed from the policy repository:
  $\mathrm{OpenPI}^{\setminus\,\mathrm{LIBERO}}$,
  $\mathrm{OpenPI}^{\setminus\,\mathrm{ALOHA}}$, and
  $\mathrm{OpenVLA}^{\setminus\,\mathrm{LIBERO}}$.
  The row groups correspond to these pruned pairs. Every cell
  reports paired values Claude Opus 4.7 / Claude Sonnet 4.6.
  The \emph{w/o template} row is omitted for both OpenPI
  repositories under both models because the upstream openpi
  codebase ships its own Tier-2 WebSocket implementation between the
  policy container and the environment endpoint
  (\S\ref{sec:method:comm}), which leaks the inter-module-transport
  prior that the \emph{w/o template} ablation is designed to remove
  from the agent's context. Sonnet 4.6 was not run under the
  \emph{w/o template} ablation for any pair (Sonnet side reported as
  ``$-$'').}
  \label{tab:ablation_per_pair}
  \setlength{\tabcolsep}{4pt}
  \renewcommand{\arraystretch}{1.05}
  \resizebox{\textwidth}{!}{%
  \begin{tabular}{l cc c c cc c}
    \toprule
    &
    \multicolumn{2}{c}{\textbf{Agent success} ($\uparrow$)}
    & \textbf{Agent inference}
    & \textbf{Interaction budget} ($\downarrow$)
    & \multicolumn{2}{c}{\textbf{Token/context budget} ($\downarrow$)}
    & \textbf{Cost} \\
    \cmidrule(lr){2-3}
    \cmidrule(lr){6-7}
    \textbf{Setting}
      & \textbf{Wrap} & \textbf{Task}
      & \textbf{Time}
      & \textbf{Agent turns}
      & \textbf{Tokens} & \textbf{Peak ctx}
      & \textbf{USD} \\
    \midrule

    \multicolumn{8}{l}{\textit{$\mathrm{OpenPI}^{\setminus\,\mathrm{LIBERO}}$}} \\
    \quad Vanilla agent       & 100.0\% / 100.0\% & 95.0\% / 90.0\%   & 21m29s / 13m47s & 199.1 / 137.0 & 15.1M / 12.4M & 136.2k / 140.7k & \$6.17 / \$5.37 \\
    \quad \sysname{}          & 100.0\% / 100.0\% & 100.0\% / 100.0\% & 26m35s / 13m27s & 185.5 / 93.0  & 17.8M / 7.6M  & 166.7k / 134.8k & \$7.75 / \$3.89 \\
    \quad w/o template        & -- / --           & -- / --           & -- / --         & -- / --       & -- / --       & -- / --         & -- / -- \\
    \quad w/o verified image  & 100.0\% / 100.0\% & 100.0\% / 75.0\%  & 29m14s / 18m53s & 416.3 / 363.0 & 16.1M / 12.8M & 145.9k / 141.3k & \$12.83 / \$11.76 \\
    \midrule

    \multicolumn{8}{l}{\textit{$\mathrm{OpenPI}^{\setminus\,\mathrm{ALOHA}}$}} \\
    \quad Vanilla agent       & 100.0\% / 100.0\% & 100.0\% / 96.0\%  & 12m53s / 8m36s  & 134.5 / 118.0 & 9.7M / 5.9M   & 111.1k / 97.8k  & \$4.07 / \$2.57 \\
    \quad \sysname{}          & 100.0\% / 100.0\% & 100.0\% / 100.0\% & 18m31s / 8m41s  & 127.0 / 57.5  & 10.9M / 4.2M  & 135.9k / 108.2k & \$5.02 / \$2.49 \\
    \quad w/o template        & -- / --           & -- / --           & -- / --         & -- / --       & -- / --       & -- / --         & -- / -- \\
    \quad w/o verified image  & -- / 100.0\%      & -- / 100.0\%      & -- / 12m27s     & -- / 212.0    & -- / 6.7M     & -- / 106.5k     & -- / \$7.74 \\
    \midrule

    \multicolumn{8}{l}{\textit{$\mathrm{OpenVLA}^{\setminus\,\mathrm{LIBERO}}$}} \\
    \quad Vanilla agent       & 100.0\% / 96.4\%  & 91.2\% / 93.6\%   & 15m08s / 20m41s & 152.0 / 219.0 & 11.9M / 19.0M & 122.1k / 151.6k & \$5.05 / \$7.62 \\
    \quad \sysname{}          & 100.0\% / 100.0\% & 96.1\% / 94.0\%   & 26m26s / 28m52s & 144.8 / 196.0 & 12.6M / 16.7M & 153.6k / 164.9k & \$6.32 / \$7.91 \\
    \quad w/o template        & 55.3\% / --       & 55.1\% / --       & 26m00s / --     & 607.5 / --    & 41.7M / --    & 219.2k / --     & \$74.94 / -- \\
    \quad w/o verified image  & 100.0\% / 100.0\% & 75.2\% / 62.6\%   & 21m22s / 49m08s & 446.0 / 615.5 & 16.4M / 27.2M & 125.0k / 150.5k & \$16.77 / \$37.64 \\

    \bottomrule
  \end{tabular}}
\end{table}

\section{Broader Impacts}
\label{app:broader-impacts}

\sysname{} is open-source research infrastructure that sits between
coding agents and the policy/benchmark/robot ecosystem. We discuss its
broader impacts under two headings: positive and negative.

\paragraph{Positive impacts.}
\sysname{} directly lowers the engineering tax of robot learning
research. Cross-family reproduction, fair evaluation, and deployment to
real hardware no longer require a fresh integration project for every
$(P, B, R)$ triple, which reduces duplicated effort across labs and
makes rigorous benchmarking more accessible to small groups that cannot
afford a full infrastructure team. Because every registry entry is
pinned to a commit hash and a container-image digest, third parties
can verify published results without trusting authors' local
configurations, which we expect to improve the reliability of the
empirical literature over time. Beyond engineering, the same
substrate---typed contracts, chambered execution, and the State
registry---is a starting point for \emph{robotic auto-research}: agents
that not only run pre-specified workflows but also propose experiments,
execute ablations, interpret results, and refine hypotheses end-to-end.
We see this autonomous-research direction as the most consequential
positive externality of the substrate, and we describe it as the
top-priority extension in Section~\ref{sec:discussion:future}.

\paragraph{Negative impacts and mitigations.}
The same automation that lowers reproduction cost also lowers the
marginal cost of deploying immature or unsafe policies on physical
hardware. An agent that can spin up a hardware container and run a
checkpoint end-to-end is, by construction, an agent that can do so for
a checkpoint that has not been adequately validated. Robot-learning
workflows further sit in a domain with non-trivial physical-safety
implications. We address these risks at the substrate level through
two mechanisms documented in the main paper. First, every workflow
stage runs inside a chambered execution container with explicit input
and output contracts, so policy-side faults cannot silently reach
hardware drivers; the simulation-pass-then-deploy ordering of
\S\ref{sec:exp:hardware} is part of this discipline. Second,
verification Sensors (\S\ref{sec:method:sensors}, Appendix~\ref{app:sensors})
gate every transition between stages on machine-checkable evidence
rather than agent self-reports. We also flag two mitigations as
planned rather than complete: human-in-the-loop modes for high-stakes
deployment, and broader hardware-side safeguards beyond the two
embodiment classes validated here. Both are discussed in
Section~\ref{sec:discussion:future} as priority follow-on work.